\documentclass[11pt]{article}

\usepackage[margin=1in]{geometry}
\usepackage[T1]{fontenc}
\usepackage{microtype}
\usepackage{amsmath,amssymb,amsthm}
\usepackage{booktabs}
\usepackage{array}
\usepackage{graphicx}
\usepackage{placeins}  
\graphicspath{{figures/}}  
\usepackage{xcolor}
\usepackage[numbers,sort&compress]{natbib}
\usepackage[colorlinks=true,linkcolor=blue,citecolor=blue,urlcolor=blue,
  hypertexnames=false]{hyperref}
\hypersetup{
  pdftitle={Causal Optimal-Transport Attention: Positional Priors and Length Extrapolation},
  pdfauthor={Ming Lian},
  pdflang={en}
}
\usepackage{thmtools,thm-restate}  

\newif\iftmlrreview

\newtheorem{theorem}{Theorem}
\newtheorem{lemma}[theorem]{Lemma}

\theoremstyle{definition}

\theoremstyle{remark}
\newtheorem{remark}[theorem]{Remark}

\newcommand{\eps}{\varepsilon}

\newcommand{\E}{\mathbb{E}}
\newcommand{\Lam}{\Lambda}
\newcommand{\astar}{{a^{*}}}  
\newcommand{\rownorm}{\operatorname{rownorm}}
\newcommand{\lse}{\operatorname{lse}}
\newcommand{\softplus}{\operatorname{softplus}}
\newcommand{\rms}{\operatorname{rms}}

\title{Causal Optimal-Transport Attention:\\
Positional Priors and Length Extrapolation}
\author{Ming Lian\\
{\normalsize Independent Researcher}\\
{\small\href{mailto:mlian@zuaa.zju.edu.cn}{\texttt{mlian@zuaa.zju.edu.cn}}}}
\date{October 31, 2026}

\begin{document}

\maketitle

\begin{abstract}
Causal masking obstructs the column balancing used by optimal-transport (OT) attention.
We study an unbalanced OT operator that uses only current and earlier rows and characterize
its induced positional prior. For bounded scores and any fixed balancing strength below
one, its fixed point assigns vanishing mass to every fixed window as context grows.
With constant scores, the prior has an exact scale-invariant form.
We classify positional encodings by whether they prevent this dilution.
We also derive finite-length window-mass bounds for log-distance priors with exponents near one.
Preregistered experiments use 25M- and 100M-parameter language models.
The OT operator improves loss extrapolation over softmax, while exploratory comparisons
favor an explicit distance bias at a fraction of the cost.
Changing only the bias beyond the training length moves long-context loss in the predicted
direction on all 30 tested checkpoints. During training, learned exponents approach one
even when constrained above it. A margin chosen from prior calculations controls
distant-mass growth in the tested setting.
\end{abstract}

\section{Introduction}
\label{sec:intro}

Softmax attention normalizes each query's scores into a distribution over keys.
In optimal-transport terms, softmax is one row-normalization step of the Sinkhorn
algorithm for an entropic OT problem. It also solves a one-sided entropic OT
problem exactly~\citep{sander2022sinkformers,litman2025sdpa}.
Iterating Sinkhorn to convergence also enforces the column marginal~\citep{sinkhorn1967concerning};
with the uniform marginal used by Sinkformers, every key then receives the same total
attention.

In causal language models, column balancing faces two obstructions
(Section~\ref{sec:background}).
First, the only lower-triangular doubly stochastic matrix is the identity, so balanced
OT degenerates.
Second, relaxing the column marginal is not enough: any column normalization whose
column sums run over all rows makes the output of row~$t$ depend on row~$t+1$, whose
query is computed from the token that row~$t$ is trained to predict.
Under teacher forcing this leaks the label and lowers the validation loss artificially.
Prefix column masses avoid this leakage~\citep{foss2026prefix}.

We address both obstructions with an unbalanced Sinkhorn iteration
whose column scaling for row~$t$ uses only the prefix mass $M_{tj}=\sum_{i\le t}P_{ij}$.
It is prefix-consistent by construction, interpolates between softmax ($\rho=0$) and a
strongly balancing regime ($\rho\to1$), and at its fixed point takes the form
\[
  \log P_{tj}=\log K_{tj}-\gamma\log M_{tj}-\lse_t,\qquad \gamma=\frac{\rho}{1-\rho},
\]
where $\lse_t$ is the log-normalizer of row $t$.
In a preregistered comparison at 100M parameters, this operator lowers the validation loss
at twice the training length by $0.45$ nats relative to softmax. It also beats log-$n$
softmax, which multiplies the scores by a factor growing like
$\log t$~\citep{chiang2022overcoming,nakanishi2025scalable}, by $0.30$.
The correction $-\gamma\log M$ enters the logits as an additive bias.
Exploratory comparisons suggest that its distance dependence accounts for the gain:
an explicit learnable log-distance bias with softmax achieves lower loss, at 40\% of the cost.

We analyze the \emph{implicit prior} of the causal OT operator: the attention
pattern it produces when content does not discriminate between keys.
The \emph{far mass} at a given context length is the attention that queries put on keys
more than the training length back. Here attention \emph{dilutes} if the mass in every
fixed-length recent window vanishes as the context grows (see Section~\ref{sec:related}
for prior terminology). A relative prior is \emph{summable} if its
weights, as a function of distance, have a finite sum; a fixed window then keeps a share
bounded away from zero, and the far mass stays bounded. The OT prior is not a function of
distance alone, so for it we speak of dilution only.

\paragraph{Contributions.}
\begin{enumerate}
  \item \textbf{Dilution.} For scores bounded by $B$, as in pre-norm Transformers, and any
  $\rho<1$, every entry of the fixed point satisfies $P_{tj}\le(e^{2B}/t)^{1-\rho}$
  (Theorem~\ref{thm:dilution}), so the mass within any fixed-length recent window vanishes as the
  context grows: the implicit prior dilutes.
  \item \textbf{Exact prior.} With constant scores,
  $tP_{t,\lfloor st\rfloor}\to(\log 1/s)^{-\rho}/\Gamma(1-\rho)$ uniformly on compact
  subsets of $(0,1)$ (Theorems~\ref{thm:lambda} and~\ref{thm:shape}): the prior depends on
  the ratio $t/j$, and near the diagonal it is an exact power law $d^{-\rho}$
  (Theorem~\ref{thm:near}). The same limit holds after any fixed number $k\ge2$ of
  iterations (Theorem~\ref{thm:iters}), and with i.i.d.\ bounded scores the fixed point
  factors into this prior times softmax-like content weights (Proposition~\ref{prop:iid}).
  \item \textbf{Far mass of positional encodings.} Writing positional schemes as potentials
  on the logits, we show that finite-dimensional group representations cannot produce a
  logarithmic kernel (Proposition~\ref{prop:rep}). We classify common encodings by whether they keep a bounded far mass or
  dilute (Proposition~\ref{prop:arch}), and give discrete near-boundary window-mass bounds
  under bounded content (Corollary~\ref{cor:boundary}).
  \item \textbf{Experiments.} The OT operator's gain over softmax holds in a preregistered
  test at 100M parameters, and exploratory analyses trace it to the distance prior
  (Section~\ref{sec:exp-ot}). Across fifteen positional schemes, far mass and long-context
  loss group by the cases of Proposition~\ref{prop:arch}, not by the form of the bias
  (Section~\ref{sec:exp-summability}). Changing a trained model's bias only beyond the
  training length moves its long-context loss in the predicted direction on all 30
  checkpoints, and a margin on a learned exponent bounds the far mass during training
  (Section~\ref{sec:exp-interventions}); CABLE beats RoPE with YaRN on every seed at three
  training lengths (Section~\ref{sec:exp-summability}).
\end{enumerate}

\paragraph{Scope.}
All theory concerns causal attention with a prefix-consistent operator.
Without a causal constraint the column marginal is a genuine constraint and none of the
results apply; we return to this in Section~\ref{sec:discussion}.
Our measure of extrapolation is the validation loss beyond the training length, which
rewards not being disturbed by distant context rather than using it.

\section{Background: attention as optimal transport under a causal mask}
\label{sec:background}

\paragraph{Attention.}
One head of a Transformer~\citep{vaswani2017attention} maps queries $q_t$, keys $k_j$ and
values $w_j$ ($t,j=1,\dots,n$; $v$ will denote a scaling vector) to outputs
$y_t=\sum_j P_{tj}w_j$ with attention weights
\begin{equation}
  P_{tj}=\frac{\exp(S_{tj}/\eps)}{\sum_{j'\in\mathcal M_t}\exp(S_{tj'}/\eps)}
  \quad(j\in\mathcal M_t),
  \qquad S_{tj}=\frac{\langle q_t,k_j\rangle}{\sqrt d}+b_{tj},
  \label{eq:softmax}
\end{equation}
where $d$ is the head dimension, $b_{tj}$ an optional additive positional bias, and
$\mathcal M_t$ the keys that query $t$ may attend to: $\mathcal M_t=\{1,\dots,t\}$ under the
causal mask of a language model, with support $\mathcal M=\{(t,j):j\le t\}$.
With the kernel $K=e^{S/\eps}$ this is $P=\rownorm(K)$, where $\rownorm$ divides each row
of $K$ by its sum over $\mathcal M_t$.
We keep the temperature $\eps$ explicit (all models use $\eps=1$) because it is the
entropic regularization strength: row $t$ of softmax maximizes
$\sum_{j\in\mathcal M_t}P_{tj}S_{tj}+\eps H(P_{t\cdot})$, $H(p)=-\sum_jp_j\log p_j$, over
probability vectors~\citep[\S\S2.2--2.3]{niculae2017regularized}.

\paragraph{Entropic optimal transport and Sinkhorn.}
Entropic OT~\citep{cuturi2013sinkhorn,peyre2019computational}, with cost $-S_{tj}$, sums
this objective over the rows and constrains both marginals,
\begin{equation}
  \max_{P\ge0}\ \sum_{t,j}P_{tj}S_{tj}+\eps H(P)
  \quad\text{s.t.}\quad P\mathbf 1=\mathbf a,\ \ P^{\top}\mathbf 1=\mathbf b,
  \label{eq:eot}
\end{equation}
with solution $P=\operatorname{diag}(u)\,K\operatorname{diag}(v)$. Sinkhorn's algorithm
alternates the row and column scalings,
\begin{equation}
  u\leftarrow \mathbf a\,/\,(Kv),\qquad v\leftarrow \mathbf b\,/\,(K^{\top}u),
  \label{eq:sinkhorn}
\end{equation}
each step enforcing one marginal exactly (divisions are entrywise).
With $\mathbf a=\mathbf 1$ and the start $v=\mathbf 1$ the first row step gives
$P=\rownorm(K)$: softmax is the first half iteration of
Sinkhorn~\citep{sander2022sinkformers}. In the log domain $\log v_j$ acts as a bias on key
$j$ that lowers the logits of keys receiving more than their share $\mathbf b_j$.

\paragraph{Unbalanced transport and $\rho$.}
Replacing the hard column constraint in \eqref{eq:eot} by a penalty
$\lambda\,\mathrm{KL}(P^{\top}\mathbf 1\,\|\,\mathbf b)$ keeps the row step and damps the
column step,
\begin{equation}
  v\leftarrow\bigl(\mathbf b\,/\,K^{\top}u\bigr)^{\rho},\qquad
  \rho=\frac{\lambda}{\lambda+\eps}\in[0,1]
  \label{eq:rho}
\end{equation}
\citep[Table~1]{chizat2018scaling}.
Only the fraction $\rho$ of the balancing bias is applied: $\rho=1$ is balanced Sinkhorn and
$\rho=0$ returns softmax.
The row constraint stays hard, so every output is a weighted average of the values.
Relaxing it too would give row $t$ a total mass growing like a power of $t$ under a causal
mask. Outputs beyond the training length would then be larger than any seen in training.

\paragraph{Two obstructions to the column constraint.}
\begin{lemma}[Triviality of causal doubly stochastic matrices]
\label{lem:identity}
A nonnegative lower-triangular matrix with all row and column sums equal to one is the
identity.
\end{lemma}
\begin{proof}
Column $n$ has a single entry $P_{nn}$, so $P_{nn}=1$ and row $n$ has no other mass.
Removing row and column $n$ leaves a matrix with the same property; induct.
\end{proof}
Relaxing the column marginal avoids this degeneracy, but column normalization over all
rows couples row $t$ to future queries. Under teacher forcing, these queries contain
ground-truth future tokens, so their contributions to the column mass leak target
information into row $t$ (Section~\ref{sec:intro}). We call an operator
\emph{prefix-consistent} if row $t$ of its output depends only on rows $1,\dots,t$ of its
input, and require it under a causal mask.

\paragraph{The prefix-consistent unbalanced operator.}
Following the prefix-mass construction of~\citet{foss2026prefix}, we replace the column mass by
$M_{tj}=\sum_{i=j}^{t}P_{ij}$ ($j\le t$), the mass that column $j$ has received from rows
$1,\dots,t$.
Starting from the all-ones array $v^{(1)}=\mathbf 1$, iteration $k$ sets
\begin{equation}
  P^{(k)}=\rownorm\bigl(K\odot v^{(k)}\bigr),\qquad
  \log v^{(k+1)}=\rho\bigl(\log v^{(k)}-\log M^{(k)}\bigr),
  \label{eq:iteration}
\end{equation}
where $M^{(k)}$ is the prefix mass of $P^{(k)}$, $\odot$ is the entrywise product, and
$\log$ acts entrywise.
Unlike the vector $v$ of~\eqref{eq:sinkhorn}, $v^{(k)}$ and $M^{(k)}$ are lower-triangular
arrays: the scaling that row $t$ applies to key $j$ depends on $t$.
Every iteration ends with a row normalization, so the output is row-stochastic; $k=1$ is
softmax.
The damping acts on the accumulated scaling $v$, as in unbalanced Sinkhorn; damping the
relative update $P\leftarrow P\odot M^{-\rho}$ instead would keep the balanced fixed point,
the identity of Lemma~\ref{lem:identity}.
At a fixed point $\log v=-\gamma\log M$ with $\gamma=\rho/(1-\rho)$:
\begin{equation}
  P_{tj}=\lambda_t\,K_{tj}\,M_{tj}^{-\gamma}\quad(j\le t),\qquad
  \sum_{j\le t}P_{tj}=1,
  \tag{FP}\label{eq:fp}
\end{equation}
where $\lambda_t>0$ is the normalizer of row $t$.
The operator solves no single OT problem: row $i<t$ is balanced against the prefix mass
available at time $i$.
The correction $-\gamma\log M_{tj}$ acts as an additive bias. Under
uniform attention $M_{tj}=\sum_{i=j}^{t}1/i\approx\log(t/j)$ grows with distance because
distant keys have received attention from more queries. The correction therefore
down-weights distant keys; Section~\ref{sec:ot-prior} derives the resulting prior.
We use $1/(1+\gamma)=1-\rho$ and $\gamma/(1+\gamma)=\rho$ throughout, and write
$\Lam_t=\sum_{i\le t}\lambda_i$ with $\Lam_0=0$; Appendix~\ref{app:notation} collects the
notation.

\paragraph{Bounded scores.}
In a pre-norm Transformer~\citep{xiong2020layer} every attention layer reads a LayerNorm
output~\citep{ba2016layer}, whose norm is bounded by a constant of the model, so queries,
keys and every quantity computed linearly from the input are bounded independently of the
length (\emph{bounded inputs}).
The scores are bounded by $B$ if $|S_{tj}/\eps|\le B$ for all $t,j$. This holds for
dot-product scores with rotary encodings~\citep{su2021roformer}, which preserve norms, and
fails for biases that decrease without bound in the distance, such as ALiBi's
$-m(t-j)$~\citep{press2022alibi} or a log-distance bias $-c\log(1+t-j)$.
Section~\ref{sec:ot-prior} shows that attention dilutes under bounded scores.
Avoiding dilution therefore requires a positional term with unbounded decay.

\section{The implicit prior of causal OT attention}
\label{sec:ot-prior}

We derive the positional prior induced by the operator without an added positional
encoding. The fixed point is computed row by row
(Section~\ref{sec:ot-recursion}); under bounded scores every fixed window loses its share
of attention (Section~\ref{sec:ot-dilution}); without content the prior has an explicit,
scale-invariant shape (Section~\ref{sec:ot-limit}). The same limit holds with finite iterations and
i.i.d.\ content (Section~\ref{sec:ot-robust}). All proofs are in Appendix~\ref{app:ot}.
Throughout, $\rho\in[0,1)$ and $P$ is the fixed point~\eqref{eq:fp} unless stated
otherwise; $L$ denotes a window length, which in the experiments is the training length.

\subsection{A row recursion}
\label{sec:ot-recursion}

Prefix consistency turns~\eqref{eq:fp} into a recursion over rows.
Given rows $1,\dots,t-1$, let $C_j=M_{t-1,j}\ge0$ be the mass that column $j$ has
received ($C_t=0$).
Since $M_{tj}=C_j+P_{tj}$, the unknowns $x_j=P_{tj}$ of row $t$ solve
\begin{equation}
  x_j\,(C_j+x_j)^{\gamma}=\lambda_tK_{tj}\quad(j\le t),\qquad\sum_jx_j=1.
  \label{eq:row}
\end{equation}
Here $C_j$ and $K_{tj}$ are known from the previous rows and the scores; only the $x_j$
and the normalizer $\lambda_t$ are unknown, which is why we drop the row index.

\begin{restatable}{lemma}{wellposed}
\label{lem:wellposed}
For any $K>0$,~\eqref{eq:fp} has a unique solution, obtained row by row from~\eqref{eq:row}.
Each $x_j$ is increasing in $\lambda_t$ and decreasing in $C_j$, and
$x_j^{1+\gamma}\le\lambda_tK_{tj}$, i.e.\ $P_{tj}^{1+\gamma}\le\lambda_tK_{tj}$, with
equality on the diagonal.
At every fixed finite length, iteration~\eqref{eq:iteration} converges to this solution.
\end{restatable}

The mass a column has received is determined, up to a constant factor, by the
normalizers of the rows that have seen it.
\begin{restatable}{lemma}{sandwich}
\label{lem:sandwich}
For $j\le t$,
\[
  \sum_{i=j}^{t}\lambda_iK_{ij}\;\le\;M_{tj}^{1+\gamma}\;\le\;(1+\gamma)\sum_{i=j}^{t}\lambda_iK_{ij}.
\]
\end{restatable}
For constant scores the sum is $\Lam_t-\Lam_{j-1}$, the clock time elapsed since key $j$
appeared. This elapsed time fixes $M_{tj}^{1+\gamma}$, and with it the correction
$-\gamma\log M_{tj}$, up to a factor between $1$ and $1+\gamma$, which tends to $1+\gamma$
away from the diagonal (Lemma~\ref{lem:ratio}). Both lemmas hold for arbitrary scores.

\subsection{Dilution}
\label{sec:ot-dilution}

To bound fixed-window mass, we first bound the row normalizer. It sets the scale of every
entry in a row and is at most of order
$1/t$:
\begin{restatable}{lemma}{lambdaupper}
\label{lem:lambda-upper}
If $|S/\eps|\le B$ then $\lambda_t\le e^{B}/t$. With constant scores, $t\lambda_t\le1$.
\end{restatable}

\begin{restatable}[Dilution under bounded scores]{theorem}{dilution}
\label{thm:dilution}
If $|S/\eps|\le B$, then for all $j\le t$
\[
  P_{tj}\le\Bigl(\frac{e^{2B}}{t}\Bigr)^{1-\rho},
  \qquad
  \sum_{t-L<j\le t}P_{tj}\le L\Bigl(\frac{e^{2B}}{t}\Bigr)^{1-\rho}\xrightarrow[t\to\infty]{}0 .
\]
\end{restatable}

For every $\rho<1$ and every bounded content, no fixed window keeps a positive share of
attention as the context grows: attention dilutes. The bound becomes informative only
when $t\gg e^{2B}$ and does not specify the attention pattern. With constant scores,
we can compute that pattern exactly.

\subsection{The scaling limit with constant scores}
\label{sec:ot-limit}

Constant scores isolate the operator's positional prior. Let
\[
  C_\rho=(1+\gamma)^{-\rho}\,\Gamma(1-\rho),\qquad
  \astar=C_\rho^{-1/(1-\rho)}=\Bigl(\frac{(1+\gamma)^{\rho}}{\Gamma(1-\rho)}\Bigr)^{1/(1-\rho)} .
\]

\begin{restatable}[Asymptotics of the row normalizers]{theorem}{rownormalizers}
\label{thm:lambda}
With constant scores, $t\lambda_t\to\astar$. For $\rho=1/2$, $\astar=2/\pi$.
\end{restatable}
The bound of Lemma~\ref{lem:lambda-upper} is thus attained up to a factor that depends on
$\rho$ alone.

\begin{restatable}[Scale-invariant limit shape of the implicit prior]{theorem}{limitshape}
\label{thm:shape}
With constant scores, for every $\delta\in(0,\frac12)$, uniformly in $s\in[\delta,1-\delta]$,
\[
  t\,P_{t,\lfloor st\rfloor}\;\longrightarrow\;f(s)=\frac{(\log 1/s)^{-\rho}}{\Gamma(1-\rho)} .
\]
\end{restatable}

The prior depends on the key only through its relative position $s=j/t$, and its total
mass is $\int_0^1f=1$. It increases toward the query, where it has an integrable
singularity at $s=1$, and decays only logarithmically toward the start of the context
($s\to0$), so a key at any fixed fraction $s$ of the context keeps a weight of order $1/t$.

\begin{figure}[t]
\centering
\includegraphics[width=\linewidth]{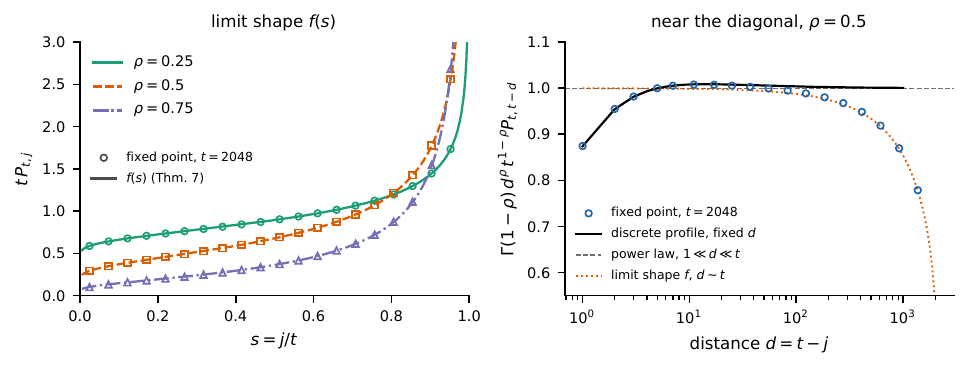}
\caption{The implicit prior with constant scores. Left: the last row of the fixed point
at $t=2048$ (markers, every 100th key) against the limit shape $f(s)$ of
Theorem~\ref{thm:shape} (lines). Right: the same row near the diagonal for $\rho=0.5$,
multiplied by $\Gamma(1-\rho)\,d^{\rho}$ so that the power law of Theorem~\ref{thm:near}(2)
is the line $1$. The fixed point (circles) follows the discrete profile of
Theorem~\ref{thm:near}(1) at small $d$, the power law for $1\ll d\ll t$, and the limit
shape $f$, here $(d/t)^{\rho}(\log\frac{t}{t-d})^{-\rho}$, as $d$ approaches $t$.}
\label{fig:prior}
\end{figure}

Theorem~\ref{thm:shape} excludes keys near the diagonal, which determine fixed-window mass.
There $\log(t/j)\approx d/t$ for $d=t-j$, and the limit becomes a power law.
\begin{restatable}[Power law near the diagonal]{theorem}{neardiagonal}
\label{thm:near}
Let $z_{-1}=0$ and let $z_d$ solve $z_d^{\gamma}(z_d-z_{d-1})=1$. With constant scores,
\begin{enumerate}
  \item for fixed $d\ge0$, $t^{1-\rho}P_{t,t-d}\to\astar^{\,1-\rho}(z_d-z_{d-1})$;
  \item for $d=d_t\to\infty$ with $d_t/t\to0$,
  $P_{t,t-d}=t^{\rho-1}d^{-\rho}\Gamma(1-\rho)^{-1}(1+o(1))$.
\end{enumerate}
\end{restatable}
The two regimes, fixed $d$ in (1) and $1\ll d\ll t$ in (2), match: by
Lemma~\ref{lem:zd}, $z_d-z_{d-1}\sim(1+\gamma)^{-\rho}d^{-\rho}$, and since
$\astar^{\,1-\rho}(1+\gamma)^{-\rho}=1/\Gamma(1-\rho)$, the limit in (1) behaves for large $d$
like the power law $d^{-\rho}/\Gamma(1-\rho)$ of (2).
Together, these limits describe the fixed point at three distance scales: a discrete profile for
$d=O(1)$, the power law $d^{-\rho}$ for $1\ll d\ll t$, and $f(s)$ for $d\sim t$
(Figure~\ref{fig:prior}).
Globally the prior depends on $t/j$, not on $t-j$: its scale invariance concerns relative
position $s=j/t$, whereas a relative position bias depends on distance alone.
For softmax every entry lies in $[e^{-2B}/t,e^{2B}/t]$; the OT operator attains the slower
rate $t^{-(1-\rho)}$ that Theorem~\ref{thm:dilution} allows: for a fixed window,
$t^{1-\rho}\sum_{t-L<j\le t}P_{tj}\to\astar^{\,1-\rho}z_{L-1}\approx L^{1-\rho}/\Gamma(2-\rho)$
(Remark~\ref{rem:window}), and for a window $L=\sigma t$ the share tends to
$\int_{1-\sigma}^1f$.

\subsection{Finite iterations and content}
\label{sec:ot-robust}

Implementations use finitely many iterations and scores that depend on content.
The following results recover the same limit shape for finite iterations with constant
scores and for the fixed point with i.i.d.\ bounded scores, respectively.
\begin{restatable}[Limit shape after finitely many iterations]{theorem}{finiteiterations}
\label{thm:iters}
With constant scores and any fixed $k\ge2$, the $k$-th iterate~\eqref{eq:iteration}
satisfies $tP^{(k)}_{t,\lfloor st\rfloor}\to f(s)$ uniformly on $[\delta,1-\delta]$.
\end{restatable}
Further iterations change only a constant, which the row normalization removes, so the
prior is set by $\rho$, not by the number of iterations. Convergence in $t$ is slow when $k$
is small and $\rho$ large: with two iterations at $t=4096$ the amplitude is $13\%$ too large
for $\rho=0.75$ (Appendix~\ref{app:num-obs}). Lemma~\ref{lem:wellposed} gives convergence
in $k$ at fixed length, without a rate uniform in length.

\begin{restatable}[Factorization of the fixed point under i.i.d.\ content]{proposition}{iidcontent}
\label{prop:iid}
Let $S_{tj}$ be i.i.d.\ with $|S/\eps|\le B$ and $\kappa=\E K$. Then, almost surely,
$t\lambda_t\to\astar/\kappa$ and, for every $\delta\in(0,\frac12)$,
\[
  \sup_{s\in[\delta,1-\delta]}\Bigl|\,t\,P_{t,\lfloor st\rfloor}
  -\frac{K_{t,\lfloor st\rfloor}}{\kappa}\,f(s)\Bigr|\to0 .
\]
\end{restatable}
Under these assumptions, the fixed point factors asymptotically into the prior $f$ and
softmax-like content weights $K/\kappa$. The column mass averages content over many rows
(numerically within $2\%$, Appendix~\ref{app:num-asymp}).

\section{A potential view of positional biases}
\label{sec:potentials}

The OT correction $-\gamma\log M_{tj}$ induces a positional prior that dilutes under
bounded scores (Section~\ref{sec:ot-prior}). Avoiding dilution requires a positional term
that decreases without bound in the distance. We now classify explicit positional
schemes by whether they keep a bounded far mass, writing each as a potential on the logits:
\[
  \log P_{tj}=\langle R(t)q_t,R(j)k_j\rangle+b_{tj}-\lse_t ,
\]
with a multiplicative group action $R$ (RoPE~\citep{su2021roformer},
GRAPE-M~\citep{zhang2026grape}), an additive or path potential $b$
(ALiBi~\citep{press2022alibi}, KERPLE~\citep{chi2022kerple},
GRAPE-A/AP, FoX~\citep{lin2025fox}, CABLE,
and the OT correction $-\gamma\log M$), and the row's log-normalizer $\lse_t$.
A bias is \emph{relative} if $b_{tj}=b(t-j)$ for all $j\le t$.
Proofs are in Appendix~\ref{app:potentials}. Removing terms that depend only on the row
simplifies several schemes:
\begin{restatable}{proposition}{rowshift}
\label{prop:rowshift}
Adding $r_t$ to row $t$ of the logits, for any $r_t$ that may depend on row $t$'s content,
leaves both softmax and the fixed point~\eqref{eq:fp} unchanged.
\end{restatable}
Consequently, a path potential whose increments do not depend on the endpoint,
$b_{tj}=\sum_{\ell=j+1}^{t}\psi(\ell)=c_t-c_j$, is the column potential $-c_j$; ALiBi and
the Forgetting Transformer are of this kind
(for ALiBi see also \citet[Prop.~G.6]{litman2026priors}).
CABLE~\citep{veisi2025cable}, $b_{tj}=g_t(S_t-S_j)$ with a content clock $S$, reduces to the
rank-one interaction $-g_tS_j$.

\subsection{Kernels of finite-dimensional representations}
Let position enter through a representation $\pi(d)=A^d$ on a space of dimension $m$:
the bias of query $t$ on key $j$ is $b_{tj}=b(t-j)$ with the kernel $b(d)=u^{\top}A^dw$,
where the vectors $u,w\in\mathbb{R}^m$ are set by the encoding's parameters and, for a
multiplicative encoding, by the content. The results below hold for every $u$ and $w$.
This covers additive encodings built from group actions and, for orthogonal $R$, the
dot-product score $\langle R(t)q,R(j)k\rangle=q^{\top}R(j-t)k$ at fixed $q,k$.
\begin{restatable}{proposition}{finiterep}
\label{prop:rep}
(i) For $d\ge m$, $b(d)=\sum_kp_k(d)\mu_k^d$ is an exponential polynomial, with the
nonzero eigenvalues $\mu_k$ of $A$ and polynomials $p_k$.
(ii) No real $b$ of this form satisfies $b(d)/\log d\to-c$ with $c\neq0$.
(iii) If the scores are bounded by $B$, softmax satisfies $P_{tj}\le e^{2B}/t$; with an
orthogonal multiplicative encoding and bounded inputs, attention therefore dilutes.
\end{restatable}
Part (ii) rules out logarithmic kernels for every finite-dimensional construction; part
(iii) says that rotary-type encodings cannot avoid dilution on their own.

\subsection{Which encodings keep a bounded far mass}
We say that attention keeps a bounded far mass if, for some window $L$, the mass beyond
distance $L$ stays below a constant smaller than one for every $t$. This excludes
dilution, where the mass beyond every fixed window tends to one. For a relative bias with
bounded content, keeping a bounded far mass is the same as having summable weights
$\sum_de^{b(d)}<\infty$: the window's share is $\sum_{d<L}e^{b(d)}$ over
$\sum_{d<t}e^{b(d)}$, up to the factor $e^{\pm2B}$. Biases that depend on content have no
single sequence of weights, so for them, and in Section~\ref{sec:exp}, we use far mass.
A linear lower bound on the decay suffices:
\begin{restatable}{lemma}{farmass}
\label{lem:farmass}
Let $\log K_{tj}=a_{tj}+b_{tj}$ with $|a_{tj}|\le B$, $b_{tt}=0$ and
$b_{tj}\le-m_0(t-j)$, $m_0>0$. For every fixed $\rho\in[0,1)$, the OT fixed point obeys
\[
  \sum_{d>L}P_{t,t-d}\le
  \frac{e^{2(1-\rho)B}e^{-(1-\rho)m_0(L+1)}}{1-e^{-(1-\rho)m_0}}
\]
for every $t$; softmax is the case $\rho=0$.
\end{restatable}
\begin{restatable}[Far mass of common encodings]{proposition}{archbounds}
\label{prop:arch}
Assume bounded inputs (Section~\ref{sec:background}). For each head:
\begin{itemize}
  \item ALiBi, FoX, GRAPE-A, GRAPE-AP and OT on ALiBi (O2) keep a bounded far mass
  for all parameter values;
  \item RoPE, GRAPE-M and the OT operator dilute for all parameter values;
  \item a log-distance bias $-c\log(1+d)$ keeps a bounded far mass if $c>1$ and dilutes if
  $c\le1$, and so does KERPLE's $-p\log(1+ad)$ with $a>0$ and $p$ in place of $c$;
  \item with an exponent $c(x_t)$ set by the query's input, on the distance (P6) or on a
  content clock (LC), the head keeps a bounded far mass if $\inf_xc(x)>1$ and dilutes if
  $\sup_xc(x)\le1$, over the admissible inputs $x$;
  \item CABLE keeps a bounded far mass if its clock's pre-activation is bounded below by a
  positive constant over the admissible inputs, and dilutes if it is never positive.
\end{itemize}
\end{restatable}
In the cases left open by the last two items, the outcome depends on the input sequence.
The relative-bias sufficiency criterion is due to~\citet{qin2024exploring};
the softmax dilution cases follow from dispersion~\citep{velickovic2025softmax}.
The bound for O2 holds at the fixed point for each fixed
$\rho<1$, and is not uniform as $\rho\to1$.

\subsection{Power priors near the boundary}
\begin{figure}[t]
\centering
\includegraphics[width=0.78\linewidth]{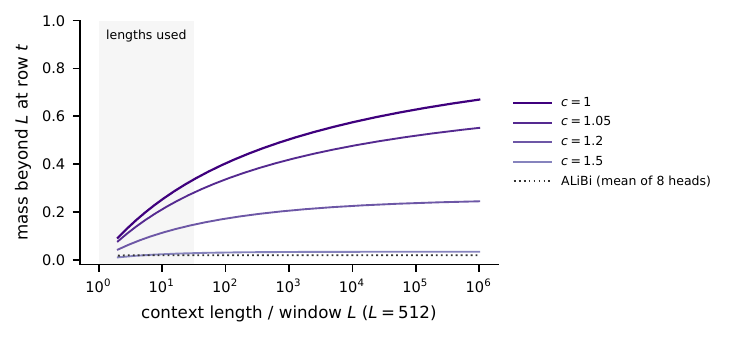}
\caption{Attention mass beyond a window $L$ at row $t$ for the prior $(1+d)^{-c}$ with
constant content (Corollary~\ref{cor:boundary}), against the context length. The shaded
range is the one used in our experiments. The linear bias $-m_hd$ (dotted) uses ALiBi's slopes
and keeps a bounded far mass
(Lemma~\ref{lem:farmass}); its far mass is averaged over eight heads.}
\label{fig:boundary}
\end{figure}

\begin{restatable}[Window mass of power-law priors]{proposition}{powerrates}
\label{prop:rates}
For the prior $(1+d)^{-c}$ and constant content, the mass within distance $L$ of the
diagonal at row $t$ is $(1-c)S_L\,t^{-(1-c)}(1+o(1))$ for $c<1$, $S_L/\log t\,(1+o(1))$
for $c=1$, and tends to $S_L/\zeta(c)>0$ for $c>1$, where
$S_L=\sum_{d<L}(1+d)^{-c}$. Bounded content changes these quantities by at most a factor
$e^{\pm2B}$.
\end{restatable}
The boundary $c=1$ dilutes only logarithmically.
For a continuous retention-window formula, see~\citet[Thm.~11.1]{marin2026predicting}.
\begin{restatable}[Log-distance exponents near one at finite length]{corollary}{boundarycor}
\label{cor:boundary}
For $c=1+\delta$ with $\delta>0$ and constant content, the mass $N_t(L)$ within distance
$L$ of the diagonal at row $t\ge\max(L,2)$ satisfies
\begin{enumerate}
  \item[(i)] $\dfrac{S_L}{1+\log t}\le N_t(L)\le\dfrac{2S_L}{\log(t+1)}$ if
  $\delta\log(t+1)\le1$;
  \item[(ii)] $\dfrac{S_L}{\zeta(c)}\le N_t(L)\le\dfrac{S_L}{\zeta(c)\,(1-t^{-\delta})}$.
\end{enumerate}
Bounded content changes these bounds by at most a factor $e^{\pm2B}$.
\end{restatable}
Up to lengths of order $e^{1/\delta}$ the prior has the logarithmic window-mass scaling of $c=1$.
For $c=1.001$ and $L=512$, even $t=32L$ gives $\delta\log(t+1)<0.01$, so this prior
stays in the logarithmic regime throughout the shaded range in Figure~\ref{fig:boundary}.
It approaches its summable limit only once $\delta\log t\gg1$.

\section{Experiments}
\label{sec:exp}

All models are decoder-only pre-norm Transformers trained from scratch on FineWeb-Edu~\citep{penedo2024fineweb}
(\texttt{sample-10BT}) with the GPT-2 tokenizer and evaluated on a held-out shard.
Extrapolation is measured by the validation loss (nats per token) beyond the training
length $L$; $L_{\mathrm{ext}}$ is its mean at $2L$ and $4L$. The far mass is the attention
on keys more than $L$ back, averaged over the queries in the second half of two validation
sequences and over layers and heads. Four experiments were preregistered, with decision
rules fixed and unit-tested before any run, and all reported runs come from committed
code; analyses outside the preregistered rules are called exploratory where they appear.
Configurations, the rules and their outcomes (Table~\ref{tab:prereg}), supplementary
results, and the public code version are in Appendix~\ref{app:exp}.
We first examine the source of the OT operator's extrapolation gain
(Section~\ref{sec:exp-ot}), then compare far mass and long-context loss across positional
schemes and training lengths (Section~\ref{sec:exp-summability}), and finally change the
tail of the bias, at test time and during training, to test its effect on loss
(Section~\ref{sec:exp-interventions}).

\subsection{The positional effect of causal OT}
\label{sec:exp-ot}

We compare the prefix-consistent operator with softmax at 99.2M non-embedding parameters,
RoPE, $L=1024$ and 2B tokens, three seeds per arm: softmax, log-$n$
softmax~\citep{chiang2022overcoming,nakanishi2025scalable}, and the operator with
$\rho\in\{0.25,0.5\}$, run for a fixed number of iterations~\eqref{eq:iteration} whose
residual is checked after training (Appendix~\ref{app:exp-setup}).

At $2L$, the loss with $\rho=0.5$ is lower than with softmax by $0.455$ and with log-$n$ softmax by $0.303$,
on every seed (Table~\ref{tab:exp0001}).

\begin{table}[!ht]
\centering
\caption{OT attention at 100M: validation loss in nats per token (mean $\pm$ sd over three
seeds) at the training length (1024) and at twice the training length. The last row is an
exploratory follow-up with the same configuration.}
\label{tab:exp0001}
\begin{tabular}{lccc}
\toprule
Arm & val@1024 & val@2048 & s/step \\
\midrule
softmax & $3.0439\pm0.0023$ & $3.5970\pm0.0285$ & 0.42 \\
log-$n$ softmax & $3.0459\pm0.0032$ & $3.4456\pm0.0596$ & 0.45 \\
OT, $\rho=0.25$ & $3.0388\pm0.0038$ & $3.3379\pm0.0471$ & 0.98 \\
OT, $\rho=0.5$ & $3.0344\pm0.0010$ & $3.1423\pm0.0544$ & 1.54 \\
\midrule
softmax + learnable $-c_h\log(1+d)$ & $\mathbf{3.0254\pm0.0023}$ & $\mathbf{3.0147\pm0.0075}$ & 0.62 \\
\bottomrule
\end{tabular}
\end{table}

Exploratory analyses support a positional explanation of this gain. In separate 25M
screening models with $\rho=0.5$, the correction $B_{tj}=-\gamma\log M_{tj}$ averaged over all
query--key pairs at the same distance $d=t-j$ decays like $-c\log(1+d)$ with $c\approx0.41$
($R^2\approx0.93$, against at most $0.75$ for a linear fit), so it multiplies the attention
weights by about $(1+d)^{-c}$; with constant scores Theorem~\ref{thm:near} gives $c=\rho$.
An explicit learnable log-distance bias with softmax beats $\rho=0.5$ at both lengths on
every seed, at 40\% of the cost (Table~\ref{tab:exp0001}, last row).
In a separate exploratory evaluation at $2L$, applying YaRN-t~\citep{peng2023yarn} to both
softmax and OT ($\rho=0.5$) shrinks their loss gap from $0.457$ to $0.011$ nats per token
(Appendix~\ref{app:exp-yarn}). The distance and content effects of the trained OT
correction remain entangled in these comparisons (Appendix~\ref{app:exp-ot}).

\subsection{Far mass and long-context loss across encodings and training lengths}
\label{sec:exp-summability}

The preregistered screening trains models with 25M non-embedding parameters, $L=512$ and
0.5B tokens, two seeds per arm, with fifteen positional schemes (Table~\ref{tab:arms}):
RoPE alone (B0), additive distance biases with linear and logarithmic kernels and learned or
input-gated exponents, a log bias on a content clock, three GRAPE
encodings~\citep{zhang2026grape}, FoX~\citep{lin2025fox}, CABLE~\citep{veisi2025cable},
and the OT operator on RoPE (O1) and on ALiBi (O2). Every scheme with a distance or path bias reaches
$L_{\mathrm{ext}}\in[3.400,3.498]$, against $3.865$ for RoPE alone and $3.573$ for OT on
RoPE. Contrary to our preregistered expectation, power-law biases do not extrapolate best:
the linear family (CABLE, ALiBi, GRAPE-A, FoX) is at least as good.
Comparing P1 (RoPE $+$ log bias) with P5 (pure log bias without RoPE) shows that unscaled
RoPE does not obstruct extrapolation when combined with a summable bias
($L_{\mathrm{ext}}=3.432$ vs $3.498$; Table~\ref{tab:screen}), while test-time windowing
removes RoPE's failure mode on B0 without retraining (Section~\ref{sec:exp-interventions}).

\begin{figure}[t]
\centering
\includegraphics[width=\linewidth]{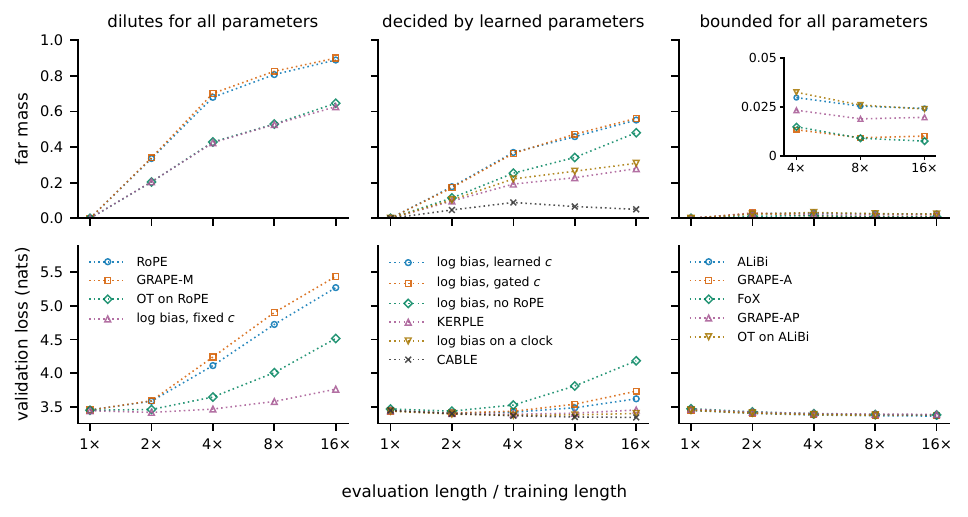}
\caption{Exploratory: far mass (top) and validation loss (bottom) against the evaluation
length at 25M, means over two seeds, one column per case of Proposition~\ref{prop:arch}.
Colours and marker shapes identify methods within each column. Each row shares its vertical axis.
The inset magnifies far mass at $4\times$--$16\times$ in the bounded case. See
Table~\ref{tab:screen} for individual values.}
\label{fig:farmass}
\end{figure}

In an exploratory analysis of the thirty checkpoints, far mass and long-context loss
vary together, with methods grouped as Proposition~\ref{prop:arch} predicts (Figure~\ref{fig:farmass}). The
schemes that dilute for all parameters gain far mass with length, up to $0.63$--$0.90$ at
$16L$, and their loss rises: the dilution of Theorem~\ref{thm:dilution} is visible already
at these lengths, although its bound is not. Those with a bounded far mass for all
parameters stay at $0.01$--$0.02$ and their loss keeps falling. Schemes whose far mass
depends on learned parameters lie in between: CABLE learns an increasing content clock and keeps its far mass at
$0.05$, and the log-bias schemes follow their learned exponents, most of which stay below
one (Table~\ref{tab:screen}). The same grouping holds at 100M
(Appendix~\ref{app:exp-confirm}).

Across encodings trained at $L=512$, higher far mass is associated with higher
long-context loss. At 100M, CABLE, the best screened scheme, also improved on RoPE with
evaluation-time YaRN scaling (B0 + YaRN-t), but fell short of the original preregistered
threshold. This threshold reflected unscaled RoPE's seed variability, which greatly
exceeded that of the paired loss differences. We therefore preregistered focused retests
with new training seeds and a threshold based on paired differences, extending the
comparison to longer training contexts (Appendix~\ref{app:exp-confirm}). CABLE
(using the learning rate selected in confirmation, $7.5\times 10^{-4}$) achieves lower
$L_{\mathrm{ext}}$ on every seed at all three training lengths, meeting the new
criterion (Table~\ref{tab:retest}). Part of this gap is already present at the training
length. Measured from each arm's own loss at $L$, the net gap beyond $L$ remains positive
and strictly above $0.01$ on every single seed ($0.028$--$0.041$; Table~\ref{tab:retest}),
confirming that extrapolation gains do not merely reflect baseline modeling improvements.
This net gap narrows with longer training contexts, alongside changes in model quality and
effective context.

\begin{table}[t]
\centering
\caption{Retest at 100M on new seeds: B0 + YaRN-t minus CABLE in $L_{\mathrm{ext}}$ (range
over seeds), B0 minus CABLE at the training length ($1\times$), and their difference, the
gap beyond $L$ measured from each arm's own loss at $L$. Absolute losses are in
Table~\ref{tab:retest-abs}.}
\label{tab:retest}
\footnotesize
\begin{tabular}{lccc}
\toprule
Training length (seeds) & gap in $L_{\mathrm{ext}}$ & gap at $1\times$ & gap beyond $L$ \\
\midrule
1024 (5) & $0.059$ ($0.055$--$0.063$) & $0.019$ & $0.040$ \\
2048 (3) & $0.065$ ($0.062$--$0.071$) & $0.030$ & $0.035$ \\
4096 (3) & $0.074$ ($0.070$--$0.078$) & $0.045$ & $0.029$ \\
\bottomrule
\end{tabular}
\end{table}

\subsection{Interventions on the tail, at test time and during training}
\label{sec:exp-interventions}

The summability criterion behind Proposition~\ref{prop:arch} motivates interventions on
the bias tail. It concerns attention dilution under bounded content scores; it does not
guarantee lower loss. We test whether removing tail decay raises long-context loss and
imposing summable decay lowers it. On trained checkpoints at 25M and 100M, we change the
total additive bias only at distances $d>L$, preserving the bias within the training
length; measured $1\times$ losses agree within $10^{-4}$. \emph{Cap} replaces $b(t,j)$ by
$b(t,t-L)$, removing tail
decay; \emph{repair} continues it as $b(t,t-L)-1.5\,[\log(1+d)-\log(1+L)]$, a summable
power-law tail;
\emph{window} subtracts $m(d-L)$ from RoPE's zero bias, with $m$ the geometric mean of the
ALiBi slopes. Caps are applied to the schemes with a bounded far mass, repairs to log-bias
schemes and windows to RoPE alone. During training, we constrain the exponents of the
learnable log bias (P1) to $c>1$, by the reparametrization $c=1+\softplus(\theta)$, and
then to $c\ge1+\delta_0$ with a margin $\delta_0$ chosen in advance from the far mass of the
prior over the lengths in use (Appendix~\ref{app:exp-train}).

\begin{figure}[t]
\centering
\includegraphics[width=\linewidth]{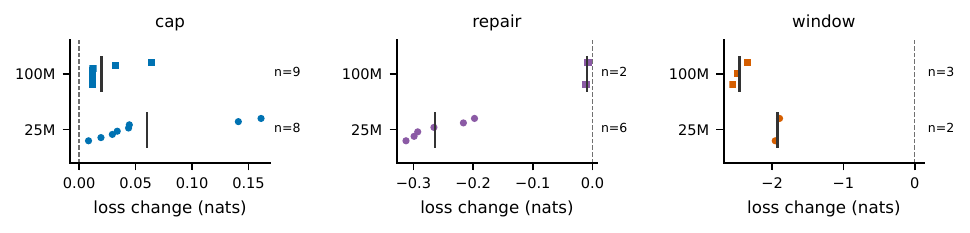}
\caption{Test-time interventions beyond the training length: loss change at the longest
length ($16\times$ at 25M, $8\times$ at 100M) for each of the 30 checkpoints. Circles and
squares mark 25M and 100M checkpoints, respectively. Dark bars show group means, and the
panels have separate linear axes. See Table~\ref{tab:tail} for values and
Figure~\ref{fig:intervention-examples} for loss curves of three checkpoints.}
\label{fig:interventions}
\end{figure}

The loss at the longest length moves in the preregistered direction on all 30
checkpoints, by $+0.039$ on average for caps, $-0.200$ for repairs and $-2.25$ for windows,
and the far mass moves with it (Figure~\ref{fig:interventions}, Table~\ref{tab:tail}). The
constraint $c>1$ lowers the $32L$ loss by $0.575$ relative to $c<1$ on every seed, but the
far mass still grows by $0.044$ from $4L$ to $16L$, above the preregistered tolerance
$0.02$, because many heads settle just above one, where by Corollary~\ref{cor:boundary}(i)
they dilute like $c=1$. With $\delta_0=0.5$ the far mass grows by $0.004$ and the $32L$
loss is $0.042$ below that of $\delta_0=0$ on every seed, at a $1\times$ loss $0.005$ higher
(Table~\ref{tab:margin}); the growth falls with the margin on every seed, consistent with
the prior-based prediction.

These test-time interventions establish that changing tail decay causally affects loss
in the models and lengths tested. They do not establish that summability generally lowers
loss, or isolate far mass from other changes, such as attention entropy
(Section~\ref{sec:discussion}); finite evaluations cannot establish asymptotic summability.
During training, the preselected margin $\delta_0=0.5$ controls far-mass growth in the
tested setting, but $\delta_0=0.25$ also meets the tolerance ($0.015$), so the smallest
sufficient margin is undetermined. At least linear decay controls far mass without an
exponent margin (Lemma~\ref{lem:farmass}); in this experiment ALiBi also has lower $32L$
loss, by $0.030$ relative to $\delta_0=0.5$.

\begin{table}[t]
\centering
\caption{Training-time margin at 25M, $L=512$: validation loss and far-mass growth from $4L$
to $16L$ (preregistered tolerance $0.02$), means over three seeds. Far mass and the share of
heads at the bound are in Appendix~\ref{app:exp-train}.}
\label{tab:margin}
\footnotesize
\begin{tabular}{lccc}
\toprule
Arm & $1\times$ & $32\times$ & far-mass growth \\
\midrule
P1, $\delta_0=0$ & 3.444 & 3.427 & 0.032 \\
P1, $\delta_0=0.25$ & 3.445 & 3.395 & 0.015 \\
P1, $\delta_0=0.5$ & 3.448 & 3.385 & 0.004 \\
ALiBi (P4) & \textbf{3.441} & \textbf{3.355} & $-0.006$ \\
\bottomrule
\end{tabular}
\end{table}

\FloatBarrier  

\section{Related work}
\label{sec:related}

\paragraph{Attention as optimal transport.}
Sinkformers~\citep{sander2022sinkformers} and efficient balancing
methods~\citep{tran2026asap,forde2026blockwise} study transport-based attention in non-causal models.
Doubly stochastic attention can slow rank collapse, although pure self-attention still
approaches rank one with depth~\citep{dong2021attention,lapenna2026rank}.
\citet{tran2026asap} list the causal column marginal as an open problem.
Sparse Sinkhorn attention~\citep{tay2020sparse}
sorts key blocks. \citet{foss2026prefix} uses prefix masses and undamped updates
$P\leftarrow\rownorm(P/M)$; we analyze damped accumulated scalings~\eqref{eq:iteration}.
GOAT learns a reference measure~\citep{litman2025sdpa,litman2026priors}; our positional prior follows from
prefix balancing. We use the damping of unbalanced Sinkhorn~\citep{chizat2018scaling,
sejourne2022unbalanced,fukunaga2022semirelaxed}. Convergence analyses for that algorithm fix matrix
size~\citep{pham2020unbalanced,sejourne2022faster}; we study length asymptotics.

\paragraph{Positional biases and length extrapolation.}
\citet{vetcha2026infinite} studies score decompositions and convergent normalization.
The relative, group-based and content-dependent schemes are described in
Section~\ref{sec:potentials}. First Passage Attention uses path extrema~\citep{liu2026firstpassage},
outside our additive-path reduction.
\citet{qin2024exploring} prove a summability condition for sliding-window approximation,
including KERPLE's threshold; we classify far mass (Proposition~\ref{prop:arch}).
\citet[Thm.~11.1]{marin2026predicting} derives a continuous retention-window formula
for finite-length power priors. The report fits effective attention exponents in
Section~9.1 and tests distance biases in Section~11.4.
We bound discrete window mass under bounded content
(Corollary~\ref{cor:boundary}), constrain explicit exponents and intervene only beyond the training length
(Section~\ref{sec:exp-interventions}). Scale-invariant attention designs distance-dependent
Gaussian logits~\citep{anson2025scaleinvariant}; our ratio prior follows from the
constant-score OT fixed point (Theorem~\ref{thm:shape}).

\paragraph{Dilution and temperature.}
Prior work uses attention dilution for diffuse linear attention~\citep{qin2022devil} and
length-induced flattening of softmax~\citep{zhang2024selective}.
\citet[Thm.~6.1]{gu2026long} proves vanishing mass on a fixed-size relevant set for
independent scores from fixed relevant and irrelevant distributions with finite
exponential moments.
Here dilution means vanishing mass in each fixed-length recent window; uniform decay of
individual attention weights implies it.
Softmax dispersion is known~\citep{velickovic2025softmax,vasylenko2026sparse};
\citet{godavarti2026accumulated} proves a far-mass lower bound under uniformly bounded
logits. Theorem~\ref{thm:dilution} extends this to the implicit prefix-mass correction.
Score scaling and sparse normalizers counteract dispersion~\citep{chiang2022overcoming,
nakanishi2025scalable,vasylenko2026sparse}; geometric limits are studied
by~\citet{mudarisov2025limitations}. Rotary selection~\citep{singh2026relative} and
data-aware rotations~\citep{levy2026darope} can improve finite-length performance.
Randomized YaRN changes training-time positions~\citep{mehta2026randomized}; our baseline
uses evaluation-time scaling. Sharp attention can still retrieve the wrong
position~\citep{spiegel2026attend}; concentration alone does not ensure retrieval.

\section{Discussion and conclusion}
\label{sec:discussion}

We characterized the positional prior of prefix-consistent causal OT, proving dilution
under bounded scores and an exact scale-invariant limit for constant scores.
We classified common positional encodings by far mass and gave discrete window-mass
bounds near the summability boundary.
Language-model experiments connect positional decay to loss extrapolation through tail
interventions and trained exponent margins, while exploratory comparisons favor explicit
distance biases as an efficient alternative to OT.

\paragraph{Theoretical limitations.}
The exact prior shape requires constant or i.i.d.\ bounded scores; multi-layer query--key
correlations and representation rank collapse in trained models~\citep{dong2021attention,lapenna2026rank}
may modify the local shape without altering the global summability bound.
The dilution bound can be loose, and we have no convergence rate for the limiting shape
or iteration bound uniform in context length.
Furthermore, the uniform-target ablation modifies fixed-point dynamics non-linearly;
a direct test-time replacement with the empirical distance-averaged profile $\bar{B}(t-j)$
would provide an appealing complementary content-free reference for future work.
The theory concerns prefix-consistent causal attention and classifies attention mass;
it gives no general loss or retrieval guarantee and does not cover bidirectional or
cross-attention.

\paragraph{Experimental limitations.}
Experiments use one corpus, 25M--100M models and limited lengths; the far-mass and retrieval
probes are exploratory, and differences in optimization and model quality may affect the comparisons.
The OT comparisons do not establish an independent benefit from column balancing, and
bias interventions change far mass together with other attention properties
(Section~\ref{sec:exp}).
The tested margin is sufficient in this setting, but may not be minimal;
low far mass and stable loss do not establish effective distant retrieval or copying
(Appendix~\ref{app:exp-distant}).

\section*{Declarations}
\paragraph{Funding and competing interests.}
This research was self-funded by the author and received no external funding.
The author declares no competing interests.

\paragraph{AI assistance.}
Anthropic Claude was used to assist with theoretical derivations, code implementation,
and experimental analysis.
OpenAI Codex assisted proof development, regression tests, literature checks and manuscript
revision, including consistency checks against code and experimental records.
The author takes responsibility for the content of this work.

\bibliographystyle{plainnat}
{\small\setlength{\bibsep}{0.5pt}\raggedright\bibliography{refs,refs-data,refs-attention}}

\appendix
\section{Notation}
\label{app:notation}

The table lists the symbols used across several sections; the last column gives the first
definition. Symbols that appear only within a proof are defined where they are used.

\begin{center}
\small
\begin{tabular}{@{}>{\raggedright\arraybackslash}p{0.12\linewidth}
  >{\raggedright\arraybackslash}p{0.69\linewidth}>{\raggedright\arraybackslash}p{0.13\linewidth}@{}}
\toprule
Symbol & Meaning & Defined in \\
\midrule
\multicolumn{3}{@{}l}{\emph{Attention and the causal OT operator}} \\
$n$ & sequence (context) length & Sec.~\ref{sec:background} \\
$t$, $j$ & query (row) and key (column) indices, $j\le t$ & Sec.~\ref{sec:background} \\
$S_{tj}$ & attention score of query $t$ for key $j$ & Eq.~\eqref{eq:softmax} \\
$b_{tj}$ & additive positional bias, part of $S_{tj}$ & Eq.~\eqref{eq:softmax} \\
$\eps$ & temperature, the entropic regularization strength & Eq.~\eqref{eq:softmax} \\
$K$ & Gibbs kernel, $K_{tj}=e^{S_{tj}/\eps}$; \emph{constant scores} means $K\equiv1$ &
  Sec.~\ref{sec:background} \\
$\mathcal M_t$ & set of keys visible to query $t$; $\{1,\dots,t\}$ under the causal mask &
  Eq.~\eqref{eq:softmax} \\
$P$ & attention matrix, row-stochastic and supported on $j\le t$; from Sec.~\ref{sec:ot-prior}
  on, the fixed point~\eqref{eq:fp} unless stated otherwise & Eq.~\eqref{eq:softmax} \\
$\rownorm$ & row normalization, dividing each row by its sum & Sec.~\ref{sec:background} \\
$\rho$ & damping of the column update, $\rho=\lambda/(\lambda+\eps)\in[0,1]$; $\rho=0$ is
  softmax, $\rho=1$ balanced Sinkhorn; all results assume $\rho<1$ & Eq.~\eqref{eq:rho} \\
$\gamma$ & exponent of the column correction, $\gamma=\rho/(1-\rho)\in[0,\infty)$ &
  Eq.~\eqref{eq:fp} \\
$M_{tj}$ & prefix column mass $\sum_{i=j}^{t}P_{ij}$, the attention key $j$ has received
  from queries $j,\dots,t$ & Sec.~\ref{sec:background} \\
$P^{(k)}$ & $k$-th iterate of the operator; $P^{(1)}$ is softmax & Eq.~\eqref{eq:iteration} \\
$\lambda_t$ & normalizer of row $t$ of the fixed point & Eq.~\eqref{eq:fp} \\
$\Lam_t$ & clock, the cumulative row normalizer: $\sum_{i\le t}\lambda_i$ & Eq.~\eqref{eq:fp} \\
$B$ & score bound, $|S_{tj}/\eps|\le B$ for all $t,j$ & Sec.~\ref{sec:background} \\
\midrule
\multicolumn{3}{@{}l}{\emph{The implicit prior}} \\
$C_\rho$ & scaling constant, $(1+\gamma)^{-\rho}\Gamma(1-\rho)$ & Sec.~\ref{sec:ot-limit} \\
$\astar$ & limit of $t\lambda_t$ with constant scores, $C_\rho^{-1/(1-\rho)}$; $2/\pi$ for
  $\rho=\frac12$ & Thm.~\ref{thm:lambda} \\
$s$ & relative position of a key, $s=j/t\in(0,1]$ & Thm.~\ref{thm:shape} \\
$f(s)$ & limit shape of the prior with constant scores, $\lim_t tP_{t,\lfloor st\rfloor}
  =(\log1/s)^{-\rho}/\Gamma(1-\rho)$ & Thm.~\ref{thm:shape} \\
$L$ & window length: the $L$ most recent keys, $t-L<j\le t$; in the experiments, the
  training length. Far mass counts the keys at $d>L$, one fewer than the complement of the
  window & Sec.~\ref{sec:ot-prior} \\
$\kappa$ & mean kernel entry, $\E K_{tj}$, for i.i.d.\ scores & Prop.~\ref{prop:iid} \\
\midrule
\multicolumn{3}{@{}l}{\emph{Positional biases}} \\
$\lse_t$ & log-normalizer of row $t$ in the potential form of the logits &
  Sec.~\ref{sec:potentials} \\
$c$ & exponent of a log-distance bias, $b_{tj}=-c\log(1+t-j)$ & Sec.~\ref{sec:potentials} \\
\bottomrule
\end{tabular}
\end{center}

\section{Proofs for Section~\ref{sec:ot-prior}}
\label{app:ot}

Notation is as in Appendix~\ref{app:notation}, with $\Lam_0=0$ and $M_{j-1,j}=0$, the
empty sum: column $j$ receives no mass before row $j$, so the first increment of every
column, on the diagonal, starts from zero ($C_t=0$ in~\eqref{eq:row}).
Small positive parameters in the proofs are written $\vartheta$ and $\nu$.

\paragraph{Structure.}
The row recursion gives well-posedness and the column-mass sandwich.
Bounds on row normalizers give dilution and feed the bootstrap of Lemma~\ref{lem:bootstrap}.
This proves Theorem~\ref{thm:lambda} and, after averaging content, Proposition~\ref{prop:iid}.
The shape results follow from Theorem~\ref{thm:lambda}; finite iterations have a separate proof.

\paragraph{Elementary facts.}
We use three, labeled by what they concern: harmonic sums, monotone sums, and an integral.
All are standard: (H) and (M) are the integral comparisons behind the asymptotics of
harmonic numbers~\citep[\S\S6.3 and~9.5]{graham1994concrete}, and (I) is Euler's integral
for the Gamma function~\citep[Eq.~5.2.1]{olver2010nist}. We include one-line proofs.
\begin{itemize}
  \item[(H)] For integers $1\le m\le t$ and $1\le a\le b$,
  $\sum_{i=m+1}^{t}1/i\le\log(t/m)$, and $\sum_{i=a}^{b}1/i$ is at least both
  $\log((b+1)/a)$ and $(b-a+1)/b$.
  The first two compare $1/i$ with $\int dx/x$ over $[i-1,i]$ and $[i,i+1]$; the third
  bounds every term by $1/b$.
  \item[(M)] Let $a<b$ be integers. If $g$ is increasing on $[a,b]$, then
  $\sum_{k=a}^{b-1}g(k)\le\int_a^bg\le\sum_{k=a+1}^{b}g(k)$, since $g(k)\le g\le g(k+1)$
  on $[k,k+1]$; if $g$ is decreasing, the inequalities reverse,
  $\sum_{k=a+1}^{b}g(k)\le\int_a^bg\le\sum_{k=a}^{b-1}g(k)$, and the left one holds even
  when $g$ has an integrable singularity at $a$. We apply it to the increasing
  $u\mapsto(\log(t/u))^{-\theta}$ and $u\mapsto(\log((t+1)/u))^{-\theta}$ with
  $\theta\in[0,1)$, and to the decreasing $u\mapsto u^{-\rho}$.
  \item[(I)] For $\theta<1$, $\int_0^1(\log 1/s)^{-\theta}\,ds=\Gamma(1-\theta)$: substitute
  $s=e^{-u}$ in $\Gamma(z)=\int_0^\infty u^{z-1}e^{-u}\,du$ with $z=1-\theta$.
\end{itemize}

\subsection{Well-posedness and the sandwich}
\label{app:ot-sandwich}

\wellposed*
\begin{proof}
Row $t$ of~\eqref{eq:fp} is~\eqref{eq:row}, which involves only rows $1,\dots,t-1$ through
$C$. For fixed $C\ge0$, $\phi_C(x)=x(C+x)^{\gamma}$ is continuous and strictly increasing
from $0$ to $\infty$ on $(0,\infty)$, so $x_j(\lambda)=\phi_{C_j}^{-1}(\lambda K_{tj})$ is
well defined, continuous, strictly increasing in $\lambda$, with limits $0$ and $\infty$.
Hence $\sum_jx_j(\lambda)$ takes the value $1$ at exactly one $\lambda_t$. Induction on $t$
gives existence and uniqueness. Since $\phi_C$ is increasing in $C$, $x_j$ decreases in
$C_j$; since $C_j\ge0$, $x^{1+\gamma}\le\phi_{C_j}(x)=\lambda_tK_{tj}$, with equality when
$C_j=0$, i.e.\ on the diagonal.

For convergence, fix row $t$. Let $K_t=(K_{t1},\dots,K_{tt})\in(0,\infty)^t$ and
$v_t^{(k)}=(v_{t1}^{(k)},\dots,v_{tt}^{(k)})\in(0,\infty)^t$ be the kernel and scaling
vectors on its visible keys, and put $z^{(k)}=\log v_t^{(k)}\in\mathbb R^t$.
For $z\in\mathbb R^t$ and a fixed prefix-mass vector $C=(C_1,\dots,C_t)\ge0$, define
$p(z)=\operatorname{softmax}(\log K_t+z)$ and $F_C(z)=\rho[z-\log(C+p(z))]$,
both in $\mathbb R^t$, with logarithms taken entrywise.
Thus $p_j(z^{(k)})=P_{tj}^{(k)}$ and
$z^{(k+1)}=F_{C^{(k)}}(z^{(k)})$, where $C_j^{(k)}=M_{t-1,j}^{(k)}$ ($C_t^{(k)}=0$).
With $\alpha_j=p_j(z)/(C_j+p_j(z))\in(0,1]$, for $1\le j,\ell\le t$,
\[
  \frac{\partial (F_C(z))_j}{\partial z_\ell}
  =\rho\bigl[(1-\alpha_j)\mathbf1_{j=\ell}+\alpha_jp_\ell(z)\bigr].
\]
These derivatives are nonnegative and sum to $\rho$ over $\ell$, so $F_C$ is a global
$\rho$-contraction in $\|\cdot\|_\infty$. Row $1$ has $P_{11}^{(k)}=1$ for all $k$.
Inductively, convergence of rows before $t$ gives $C^{(k)}\to C^*$.
Let $z^*$ be the unique fixed point of $F_{C^*}$. Then
\[
  \|z^{(k+1)}-z^*\|_\infty\le\rho\|z^{(k)}-z^*\|_\infty+\eta_k,
  \qquad \eta_k=\|F_{C^{(k)}}(z^*)-F_{C^*}(z^*)\|_\infty\to0.
\]
For $\rho<1$ this implies $z^{(k)}\to z^*$: iterate the inequality and split its
geometrically weighted error sum at any fixed index. Induction completes the proof;
the limit satisfies~\eqref{eq:fp}. No rate uniform in $t$ is asserted.

\end{proof}

\sandwich*
\begin{proof}
Fix column $j$. Row $i\ge j$ adds $x=P_{ij}$ to $C=M_{i-1,j}$, with
$x(C+x)^{\gamma}=\lambda_iK_{ij}$. Then
\begin{equation}
  (C+x)^{1+\gamma}-C^{1+\gamma}=(1+\gamma)\int_C^{C+x}u^{\gamma}\,du
  \in\bigl[(1+\gamma)xC^{\gamma},\,(1+\gamma)x(C+x)^{\gamma}\bigr],
  \label{eq:increment}
\end{equation}
and the right end equals $(1+\gamma)\lambda_iK_{ij}$. Summing over $i=j,\dots,t$ gives the
upper bound. For the lower bound, column mass never decreases, so
$P_{ij}=\lambda_iK_{ij}M_{ij}^{-\gamma}\ge\lambda_iK_{ij}M_{tj}^{-\gamma}$ for $i\le t$;
summing, $M_{tj}\ge M_{tj}^{-\gamma}\sum_{i=j}^{t}\lambda_iK_{ij}$.
\end{proof}

The left end of~\eqref{eq:increment} gives a second lower bound, used in
Lemma~\ref{lem:ratio}, which is nearly tight where the increments are small against the
accumulated mass.
Since $M_{j-1,j}=0$, the column mass telescopes over the rows that see it,
\[
  M_{tj}^{1+\gamma}=\sum_{i=j}^{t}\bigl(M_{ij}^{1+\gamma}-M_{i-1,j}^{1+\gamma}\bigr),
\]
and the term of row $i$ is the increment~\eqref{eq:increment} with $C=M_{i-1,j}$ and
$x=P_{ij}$.
On the diagonal $C=0$, and the term is $P_{jj}^{1+\gamma}=\lambda_jK_{jj}$ by
Lemma~\ref{lem:wellposed}.
For $i>j$, write the increment ratio $r_i=P_{ij}/M_{ij}\in(0,1]$, so that
$C=M_{ij}-P_{ij}=(1-r_i)M_{ij}$; the left end of~\eqref{eq:increment} and
$P_{ij}M_{ij}^{\gamma}=\lambda_iK_{ij}$ from~\eqref{eq:fp} give
\[
  M_{ij}^{1+\gamma}-M_{i-1,j}^{1+\gamma}\ge(1+\gamma)P_{ij}\bigl((1-r_i)M_{ij}\bigr)^{\gamma}
  =(1+\gamma)(1-r_i)^{\gamma}\lambda_iK_{ij}.
\]
Summing,
\begin{equation}
  M_{tj}^{1+\gamma}\ge\lambda_jK_{jj}+(1+\gamma)\sum_{i=j+1}^{t}\lambda_iK_{ij}(1-r_i)^{\gamma}.
  \label{eq:sharp-lower}
\end{equation}
Each term differs from the upper bound of Lemma~\ref{lem:sandwich} only by the factor
$(1-r_i)^{\gamma}$, which tends to one as $r_i\to0$; for $r_i$ near one,~\eqref{eq:sharp-lower}
is weaker than the lower bound of Lemma~\ref{lem:sandwich}.

\subsection{Bounded scores: dilution, the a priori bound, and the increment ratio}
\label{app:ot-bounded}

Throughout this subsection the scores are bounded, $|S/\eps|\le B$, that is
$K_{tj}\in[e^{-B},e^{B}]$ for all $t,j$; constant scores are the case $B=0$, $K\equiv1$
(Proposition~\ref{prop:rowshift}), and every constant below is then to be read at $B=0$.
We first bound the row normalizers from above, which gives dilution, and then from below,
together with the ratio of a row's increment to a column's mass; the last two results
feed the proof of Theorem~\ref{thm:lambda} and of Proposition~\ref{prop:iid}.

\lambdaupper*
\begin{proof}
Rows $1,\dots,t$ carry total mass $t$ on columns $1,\dots,t$, so $\sum_{j\le t}M_{tj}=t$.
By~\eqref{eq:fp} and Jensen's inequality for the convex map $u\mapsto u^{-\gamma}$,
\[
  1=\lambda_t\sum_{j\le t}K_{tj}M_{tj}^{-\gamma}
  \ge\lambda_te^{-B}\sum_{j\le t}M_{tj}^{-\gamma}
  \ge\lambda_te^{-B}\,t\,\Bigl(\frac1t\sum_{j\le t}M_{tj}\Bigr)^{-\gamma}
  =\lambda_te^{-B}t .
  \qedhere
\]
\end{proof}

\dilution*
\begin{proof}
By Lemmas~\ref{lem:wellposed} and~\ref{lem:lambda-upper},
$P_{tj}^{1+\gamma}\le\lambda_tK_{tj}\le e^{B}\cdot e^{B}/t$; take the power
$1/(1+\gamma)=1-\rho$. A window of length $L$ has at most $L$ entries.
\end{proof}

\paragraph{The column sum and the normalization.} Write
\[
  W_{tj}=\sum_{i=j}^{t}\lambda_iK_{ij},\qquad\text{so that}\qquad
  W_{tj}\le M_{tj}^{1+\gamma}\le(1+\gamma)W_{tj}
\]
by Lemma~\ref{lem:sandwich}, and $e^{-B}(\Lam_t-\Lam_{j-1})\le W_{tj}\le e^{B}(\Lam_t-\Lam_{j-1})$;
with constant scores $W_{tj}=\Lam_t-\Lam_{j-1}$. Lemma~\ref{lem:sandwich} and
$-\gamma/(1+\gamma)=-\rho$ turn the row normalization
$1/\lambda_t=\sum_{j\le t}K_{tj}M_{tj}^{-\gamma}$ into
\begin{equation}
  e^{-B}\sum_{j\le t}\bigl((1+\gamma)W_{tj}\bigr)^{-\rho}
  \;\le\;\frac1{\lambda_t}\;\le\;e^{B}\sum_{j\le t}W_{tj}^{-\rho}.
  \tag{S$_B$}\label{eq:SB}
\end{equation}
The next two lemmas prepare the proofs of Theorem~\ref{thm:lambda}, which explains how
they are used, and of Proposition~\ref{prop:iid}: an a priori lower bound on $t\lambda_t$
from the right side of~\eqref{eq:SB}, and a bound on the ratio of a row's increment to a
column's mass.

\begin{lemma}
\label{lem:alpha0}
If $K_{tj}\in[e^{-B},e^{B}]$ for all $t,j$, then $t\lambda_t\ge a_B$ for all $t$, where
$a_B^{1-\rho}=e^{-(1+\rho)B}/(2\Gamma(1-\rho))$. For constant scores this is
$t\lambda_t\ge a_0$ with $a_0^{1-\rho}=1/(2\Gamma(1-\rho))$.
\end{lemma}
\begin{proof}
Induction on $t$. Row $1$ has the single entry $P_{11}=1$, so $\lambda_1=1/K_{11}\ge e^{-B}
\ge a_B$, since $a_B^{1-\rho}\le e^{-(1+\rho)B}\le e^{-(1-\rho)B}$. Assume
$\lambda_i\ge a_B/i$ for $i<t$. For $m\le t-2$, (H) gives
$\Lam_t-\Lam_m\ge a_B\sum_{i=m+1}^{t-1}1/i\ge a_B\log(t/(m+1))$, and
$\Lam_t-\Lam_{t-1}=\lambda_t$. The right side of~\eqref{eq:SB} with
$W_{tj}\ge e^{-B}(\Lam_t-\Lam_{j-1})$, then (M) and (I), give
\[
  \frac1{\lambda_t}\le e^{(1+\rho)B}\Bigl(\lambda_t^{-\rho}
  +a_B^{-\rho}\sum_{k=1}^{t-1}(\log(t/k))^{-\rho}\Bigr)
  \le e^{(1+\rho)B}\bigl(\lambda_t^{-\rho}+a_B^{-\rho}\,t\,\Gamma(1-\rho)\bigr).
\]
With $y=t\lambda_t$ this reads
$1\le e^{(1+\rho)B}\bigl((y/t)^{1-\rho}+y\,a_B^{-\rho}\Gamma(1-\rho)\bigr)$. If $y<a_B$, the
right side is below $e^{(1+\rho)B}a_B^{1-\rho}(1+\Gamma(1-\rho))\le
2e^{(1+\rho)B}\Gamma(1-\rho)\,a_B^{1-\rho}=1$, because $\Gamma(1-\rho)\ge1$;
contradiction.
\end{proof}

\begin{lemma}
\label{lem:ratio}
Let $K_{tj}\in[e^{-B},e^{B}]$ for all $t,j$ and $c_B=e^{3B}/a_B$. For $i\ge j$,
\[
  r_i=\frac{P_{ij}}{M_{ij}}\le\frac{\lambda_iK_{ij}}{W_{ij}}\le\frac{c_B}{i-j+1};
\]
for constant scores this is $r_i\le\lambda_i/(\Lam_i-\Lam_{j-1})\le1/(a_0(i-j+1))$.
Consequently the relative deficit
\[
  \eta_{tj}=1-\frac{M_{tj}^{1+\gamma}}{(1+\gamma)W_{tj}}\in[0,1)
\]
is small away from the diagonal. There are constants $C_1,C_2$ depending only on $\rho$
and $B$ such that for all $t\ge1$:
\begin{enumerate}
  \item for every $\vartheta\in(0,\frac12)$,
  $\displaystyle0\le\max_{\vartheta t\le j\le(1-\vartheta)t}\eta_{tj}
  \le\frac{C_1\log(t+1)}{\vartheta t\log\frac1{1-\vartheta}}$, which tends to $0$;
  \item for $t/2\le j\le t$ and $d=t-j$, $\displaystyle0\le\eta_{tj}\le
  C_2\,\frac{1+\log(d+1)}{d+1}$.
\end{enumerate}
These are upper bounds; only their convergence to zero is used below.
\end{lemma}
\begin{proof}
\emph{Ratio.} We use only the lower bound of Lemma~\ref{lem:sandwich},
$M_{ij}^{1+\gamma}\ge W_{ij}$. Raised to the decreasing power $-\rho=-\gamma/(1+\gamma)$
and to the increasing power $1-\rho=1/(1+\gamma)$, it bounds the numerator and the
denominator of $r_i$:
\[
  P_{ij}=\lambda_iK_{ij}M_{ij}^{-\gamma}=\lambda_iK_{ij}\bigl(M_{ij}^{1+\gamma}\bigr)^{-\rho}
  \le\lambda_iK_{ij}W_{ij}^{-\rho},\qquad
  M_{ij}=\bigl(M_{ij}^{1+\gamma}\bigr)^{1-\rho}\ge W_{ij}^{1-\rho};
\]
dividing gives $r_i\le\lambda_iK_{ij}/W_{ij}$. By Lemma~\ref{lem:lambda-upper},
$\lambda_iK_{ij}\le e^{2B}/i$; by Lemma~\ref{lem:alpha0} and (H),
$W_{ij}\ge e^{-B}(\Lam_i-\Lam_{j-1})\ge e^{-B}a_B\sum_{k=j}^{i}1/k\ge e^{-B}a_B(i-j+1)/i$.

\emph{Deficit.} $\eta_{tj}\ge0$ by Lemma~\ref{lem:sandwich}. For the upper bound, compare
\eqref{eq:sharp-lower} with $(1+\gamma)W_{tj}=(1+\gamma)\sum_{i=j}^{t}\lambda_iK_{ij}$ and
use $(1-r)^{\gamma}\ge1-\max(1,\gamma)r$ on $[0,1]$: with
$c_\gamma=(1+\gamma)\max(1,\gamma)$, the deficit satisfies
\[
  (1+\gamma)W_{tj}-M_{tj}^{1+\gamma}
  \le\gamma\lambda_jK_{jj}+c_\gamma\sum_{i=j+1}^{t}\lambda_iK_{ij}r_i .
\]
It remains to bound this right side, with the ratio bound just proved, and the
denominator of $\eta$ in each range.

\emph{Case (1).} For $i\ge j$, $\lambda_iK_{ij}\le e^{2B}/i\le e^{2B}/j$, which takes out a
common factor. With the ratio bound and $k=i-j$,
\[
  \sum_{i=j+1}^{t}\lambda_iK_{ij}r_i\le\frac{e^{2B}c_B}{j}\sum_{k=1}^{t-j}\frac1{k+1}
  =\frac{e^{2B}c_B}{j}\sum_{q=2}^{t-j+1}\frac1q\le\frac{e^{2B}c_B\log(t+1)}{j},
\]
extending the sum to $q\le t+1$ and applying the first bound of (H) with $m=1$. Together
with $\gamma\lambda_jK_{jj}\le\gamma e^{2B}/j$, the deficit is at most
$e^{2B}j^{-1}\bigl(\gamma+c_\gamma c_B\log(t+1)\bigr)$ for every $j$. For
$j\le(1-\vartheta)t$, Lemma~\ref{lem:alpha0} and (H) bound the denominator of $\eta$ below
by a constant,
\[
  (1+\gamma)W_{tj}\ge(1+\gamma)e^{-B}a_B\log\frac{t+1}{j}
  \ge(1+\gamma)e^{-B}a_B\log\frac1{1-\vartheta}.
\]
For $j\ge\vartheta t$ the deficit is at most
$e^{2B}(\vartheta t)^{-1}(\gamma+c_\gamma c_B\log(t+1))$; since
$\gamma\le\gamma\log(t+1)/\log2$, dividing gives (1) with
$C_1=e^{3B}(\gamma/\log2+c_\gamma c_B)/((1+\gamma)a_B)$, which does not depend on
$\vartheta$.

\emph{Case (2).} For $j\ge t/2$, $\lambda_iK_{ij}\le e^{2B}/i\le2e^{2B}/t$, and as in
case (1), now with $t-j=d$,
\[
  \sum_{i=j+1}^{t}\lambda_iK_{ij}r_i\le\frac{2e^{2B}c_B}{t}\sum_{k=1}^{d}\frac1{k+1}
  \le\frac{2e^{2B}c_B\log(d+1)}{t},
\]
so the deficit is at most $\frac{2e^{2B}}t\bigl(\gamma+c_\gamma c_B\log(d+1)\bigr)$. By (H)
the denominator is at least
\[
  (1+\gamma)e^{-B}a_B\sum_{k=j}^{t}\frac1k\ge\frac{(1+\gamma)e^{-B}a_B(d+1)}t .
\]
Dividing gives (2) with $C_2=2e^{3B}\max(\gamma,c_\gamma c_B)/((1+\gamma)a_B)$.
For $B=0$, $c_0=1/a_0$ and these constants reduce to those of the constant-score case.
\end{proof}

\subsection{Auxiliary estimates: the ends of a row and a bootstrap}
\label{app:ot-bootstrap}

Two arguments recur in the proofs of Theorems~\ref{thm:lambda} and~\ref{thm:iters} and
Proposition~\ref{prop:iid}, and we state them once.

\emph{The ends of a row.} Write $\ell_{tj}=\log\frac{t+1}j$ for $1\le j\le t$. Each of
these proofs bounds a row by the profile $\ell_{tj}^{-\rho}$, and needs that the keys far
from the query ($j<\vartheta t$) and near it ($j>(1-\vartheta)t$) carry a share that
vanishes with $\vartheta$, uniformly in $t$.
\begin{lemma}
\label{lem:row-ends}
Let $v_{tj}\ge0$ with $v_{tj}\le C\,\ell_{tj}^{-\rho}$ for all $1\le j\le t$. Then for
$\vartheta\in(0,\frac12)$ and $t\ge1/\vartheta$,
\[
  \frac1t\sum_{j<\vartheta t}v_{tj}\le C\vartheta\Bigl(\log\frac1\vartheta\Bigr)^{-\rho},
  \qquad
  \frac1t\sum_{j>(1-\vartheta)t}v_{tj}\le\frac{2C\vartheta^{1-\rho}}{1-\rho}.
\]
\end{lemma}
\begin{proof}
For $j<\vartheta t$, $\ell_{tj}\ge\log\frac1\vartheta$, and there are fewer than
$\vartheta t$ such $j$. For $j>(1-\vartheta)t$, $\log x\ge1-1/x$ gives
$\ell_{tj}\ge r/(t+1)$ with $r=t+1-j\le\vartheta t+1\le2\vartheta t$, and (M) for the
decreasing $u\mapsto u^{-\rho}$ gives
\[
  \frac1t\sum_{j>(1-\vartheta)t}\ell_{tj}^{-\rho}\le\frac{(t+1)^{\rho}}t
  \sum_{r\le2\vartheta t}r^{-\rho}\le\frac{(2t)^{\rho}}t\,\frac{(2\vartheta t)^{1-\rho}}{1-\rho}
  =\frac{2\vartheta^{1-\rho}}{1-\rho}.\qedhere
\]
\end{proof}

\emph{A bootstrap.} The proofs of Theorem~\ref{thm:lambda} and Proposition~\ref{prop:iid}
share one argument,
which we isolate here as a statement about a positive sequence $\lambda'_t$: the row
normalizers, rescaled by $\kappa=\E K$ in Proposition~\ref{prop:iid}. Write
$\Lam'_t=\sum_{i\le t}\lambda'_i$, $\Lam'_0=0$, $D_{tj}=\Lam'_t-\Lam'_{j-1}$,
\[
  J_t(\vartheta)=\{j:\vartheta t\le j\le(1-\vartheta)t\},\qquad
  \chi(\vartheta)=\vartheta^{1-\rho}+\vartheta\Bigl(\log\frac1\vartheta\Bigr)^{-\rho},\qquad
  T(x)=\frac{x^{\rho}}{C_\rho}.
\]
The row normalization expresses $1/\lambda'_t$ as a sum over the differences $D_{tj}$,
which are built from the earlier normalizers, so an asymptotic bound $i\lambda'_i\le x$ (or
$\ge x$) on the earlier rows yields a bound on the current one. The lemma assumes the two
inequalities that make this work: in the middle range $J_t(\vartheta)$, a lower and an
upper bound on $1/\lambda'_t$ carrying the same factor $(1+\gamma)^{-\rho}$, up to errors
that vanish with $\nu$ and $\vartheta$. Both then turn a bound $x$ on the earlier rows into
the bound $T(x)$ on the current row. Since $\log T(x)=\rho\log x-\log C_\rho$ has slope
$\rho<1$, $T$ is a contraction in log scale; $\astar=C_\rho^{-1/(1-\rho)}$ is its unique
fixed point on $(0,\infty)$, and $x\le T(x)$ if and only if $x\le\astar$. Applied to the
limits superior and inferior, this pins both to $\astar$.

\begin{lemma}
\label{lem:bootstrap}
Let $\lambda'_t>0$ with $0<\liminf t\lambda'_t\le\limsup t\lambda'_t<\infty$, and suppose
there is a constant $C$ such that for every $\vartheta,\nu\in(0,\frac12)$,
for all large $t$,
\begin{enumerate}
  \item[(i)] $\displaystyle\frac1{\lambda'_t}\ge\sum_{j\in J_t(\vartheta)}
  \bigl((1+\gamma)(1+\nu)D_{tj}\bigr)^{-\rho}-\nu t$;
  \item[(ii)] $\displaystyle\frac1{\lambda'_t}\le\sum_{j\in J_t(\vartheta)}
  \bigl((1+\gamma)(1-\nu)D_{tj}\bigr)^{-\rho}+t\bigl(\nu+C\chi(\vartheta)\bigr)$.
\end{enumerate}
Then $t\lambda'_t\to\astar$.
\end{lemma}
\begin{proof}
Let $\overline a=\limsup t\lambda'_t$ and $\underline a=\liminf t\lambda'_t$, and
$I_\vartheta=\int_\vartheta^{1-\vartheta}(\log\frac1s)^{-\rho}ds$, which tends to
$\Gamma(1-\rho)$ as $\vartheta\to0$ by (I).

\emph{Upper bound: $\overline a\le\astar$.} Fix $\vartheta$ and $\nu$, and $i_0$ with
$\lambda'_i\le(\overline a+\nu)/i$ for $i\ge i_0$. For large $t$, every $j\in J_t(\vartheta)$
satisfies $j-1\ge i_0$, so (H) gives $D_{tj}\le(\overline a+\nu)\log\frac t{j-1}$. Since
$u\mapsto(\log\frac tu)^{-\rho}$ is increasing, (M) and the substitution $u=ts$ give
$\sum_{j\in J_t(\vartheta)}(\log\frac t{j-1})^{-\rho}\ge t\,I_\vartheta(1+o(1))$, and (i)
yields
\[
  \frac1{\overline a}=\liminf_{t\to\infty}\frac1{t\lambda'_t}
  \ge\bigl((1+\gamma)(1+\nu)(\overline a+\nu)\bigr)^{-\rho}I_\vartheta-\nu .
\]
Letting $\nu\to0$ and then $\vartheta\to0$ gives
$1/\overline a\ge((1+\gamma)\overline a)^{-\rho}\Gamma(1-\rho)=C_\rho\,\overline a^{-\rho}$,
that is, $\overline a\le T(\overline a)$, so $\overline a\le\astar$.

\emph{Lower bound: $\underline a\ge\astar$.} Fix $\vartheta$, $\nu<\underline a$, and $i_1$
with $\lambda'_i\ge(\underline a-\nu)/i$ for $i\ge i_1$. For large $t$, (H) gives
$D_{tj}\ge(\underline a-\nu)\log\frac{t+1}j$ for $j\in J_t(\vartheta)$, and by (M) and (I),
$\sum_{j\le t}(\log\frac{t+1}j)^{-\rho}\le(t+1)\Gamma(1-\rho)$. So (ii) yields
\[
  \frac1{\underline a}=\limsup_{t\to\infty}\frac1{t\lambda'_t}
  \le\bigl((1+\gamma)(1-\nu)(\underline a-\nu)\bigr)^{-\rho}\Gamma(1-\rho)
  +\nu+C\chi(\vartheta).
\]
Letting $\nu\to0$ and then $\vartheta\to0$ gives $1/\underline a\le C_\rho\,\underline a^{-\rho}$,
that is, $\underline a\ge T(\underline a)$, so $\underline a\ge\astar$. Since
$\underline a\le\overline a$, both equal $\astar$.
\end{proof}

\subsection{Proof of Theorem~\ref{thm:lambda}}
\label{app:ot-lambda}

\rownormalizers*

From here on, until Proposition~\ref{prop:iid}, the scores are constant: $K\equiv1$, so
$B=0$, $W_{tj}=\Lam_t-\Lam_{j-1}$, and Lemmas~\ref{lem:alpha0} and~\ref{lem:ratio} hold
with $a_0$ and $c_0=1/a_0$. With $m=j-1$, \eqref{eq:SB} becomes the
\emph{normalization sandwich}
\begin{equation}
  \sum_{m=0}^{t-1}\bigl((1+\gamma)(\Lam_t-\Lam_m)\bigr)^{-\rho}
  \;\le\;\frac1{\lambda_t}\;\le\;\sum_{m=0}^{t-1}(\Lam_t-\Lam_m)^{-\rho},
  \tag{S}\label{eq:S}
\end{equation}
whose two sides differ by the factor $(1+\gamma)^{-\rho}$.

\emph{Strategy.} We apply Lemma~\ref{lem:bootstrap} with $\lambda'=\lambda$, so that
$D_{tj}=\Lam_t-\Lam_{j-1}$. Hypothesis (i) is the left side of~\eqref{eq:S}. For (ii), the
right side of~\eqref{eq:S} alone would not do: it lacks the factor $(1+\gamma)^{-\rho}$,
and the argument of the lemma would give only
$\liminf t\lambda_t\ge\Gamma(1-\rho)^{-1/(1-\rho)}$, which lies below $\astar$. Away from
the diagonal, however, a row's increment is negligible against the mass the column has
accumulated, and Lemma~\ref{lem:ratio}(1) shows that $M_{tj}^{1+\gamma}$ is then close to
the upper end of Lemma~\ref{lem:sandwich}. This restores the factor in the middle range;
the rest of the sum is the error $C\chi(\vartheta)$.

\begin{proof}
By Lemmas~\ref{lem:lambda-upper} and~\ref{lem:alpha0}, $a_0\le t\lambda_t\le1$. Fix
$\vartheta,\nu\in(0,\frac12)$.

\emph{Hypothesis (i).} The terms on the left of~\eqref{eq:S} are positive, so keeping only
$j=m+1\in J_t(\vartheta)$ gives
$1/\lambda_t\ge\sum_{j\in J_t(\vartheta)}((1+\gamma)D_{tj})^{-\rho}$, which is at least
the right side of (i).

\emph{Hypothesis (ii).} Split the normalization
$1/\lambda_t=\sum_{j\le t}M_{tj}^{-\gamma}$ into the middle range $J_t(\vartheta)$ and the
two ends. By Lemma~\ref{lem:sandwich}, Lemma~\ref{lem:alpha0} and (H),
\begin{equation}
  M_{tj}^{-\gamma}=\bigl(M_{tj}^{1+\gamma}\bigr)^{-\rho}\le(\Lam_t-\Lam_{j-1})^{-\rho}
  \le a_0^{-\rho}\ell_{tj}^{-\rho},
  \label{eq:const-profile}
\end{equation}
so by Lemma~\ref{lem:row-ends} the two ends contribute at most $t\,C\chi(\vartheta)$ with
$C=2a_0^{-\rho}/(1-\rho)$. In the middle range, Lemma~\ref{lem:ratio}(1) gives
$M_{tj}^{1+\gamma}\ge(1+\gamma)(1-\bar\eta_t)D_{tj}$ with $\bar\eta_t\to0$, so for large
$t$ this part is at most $\sum_{j\in J_t(\vartheta)}((1+\gamma)(1-\nu)D_{tj})^{-\rho}$.
Together these give (ii).

Lemma~\ref{lem:bootstrap} gives $t\lambda_t\to\astar$. Finally, for $\rho=\frac12$,
$\gamma=1$, $C_\rho=2^{-1/2}\Gamma(\frac12)=2^{-1/2}\sqrt\pi$, and
$\astar=C_\rho^{-1/(1-\rho)}=C_\rho^{-2}=2/\pi$.
\end{proof}

\subsection{Proofs of Theorems~\ref{thm:shape} and~\ref{thm:near}}
\label{app:ot-shape}

\limitshape*
\begin{proof}
$P_{tj}=\lambda_tM_{tj}^{-\gamma}$. Take $\vartheta=\delta$ in Lemma~\ref{lem:ratio}(1):
$M_{tj}^{1+\gamma}=(1+\gamma)(\Lam_t-\Lam_{j-1})(1+o(1))$ uniformly for
$j=\lfloor st\rfloor$, $s\in[\delta,1-\delta]$.
Theorem~\ref{thm:lambda} says that $\omega_n=\sup_{i\ge n}|i\lambda_i-\astar|\to0$. Every
$i$ in $\Lam_t-\Lam_{j-1}=\sum_{i=j}^{t}\lambda_i$ satisfies $i\ge j\ge\delta t-1$, so
\[
  \Bigl|\Lam_t-\Lam_{j-1}-\astar\sum_{i=j}^{t}\frac1i\Bigr|
  \le\omega_{\lceil\delta t\rceil-1}\sum_{i=j}^{t}\frac1i
  \le\omega_{\lceil\delta t\rceil-1}\Bigl(\log\frac1\delta+1\Bigr)\to0
\]
for $t$ large, with a bound that does not depend on $s$; with (H),
$\sum_{i=j}^{t}1/i=\log(1/s)+O(1/(\delta t))$. Hence
$\Lam_t-\Lam_{j-1}=\astar\log(1/s)+o(1)$ uniformly in $s\in[\delta,1-\delta]$, and since
$\astar\log(1/s)\ge\astar\log\frac1{1-\delta}>0$ there, the additive $o(1)$ is a factor
$1+o(1)$. Hence
\[
  tP_{tj}=t\lambda_t\bigl((1+\gamma)\astar\log(1/s)\bigr)^{-\rho}(1+o(1))
  \to\astar\bigl((1+\gamma)\astar\bigr)^{-\rho}(\log1/s)^{-\rho},
\]
and $\astar^{\,1-\rho}(1+\gamma)^{-\rho}=1/\Gamma(1-\rho)$ by the definition of $\astar$.
\end{proof}

Near the diagonal the row normalizers vary little across the rows that a column has seen,
so the column behaves as if every normalizer were equal. The discrete profile $z_d$ of
Theorem~\ref{thm:near} is that column with all normalizers equal to one.
\begin{lemma}
\label{lem:zd}
Let $z_{-1}=0$ and $z_d^{\gamma}(z_d-z_{d-1})=1$. Then $z_d$ is the mass of a column after
$d+1$ rows when every row normalizer equals one, $z_d^{1+\gamma}=(1+\gamma)(d+1)(1+o(1))$,
and $z_d-z_{d-1}\sim(1+\gamma)^{-\rho}d^{-\rho}$.
\end{lemma}
\begin{proof}
Each increment is $z_d-z_{d-1}=z_d^{-\gamma}$, which is~\eqref{eq:row} for one column
with $\lambda=K=1$. The argument of Lemma~\ref{lem:sandwich} with $\lambda\equiv1$ gives
$d+1\le z_d^{1+\gamma}\le(1+\gamma)(d+1)$, and~\eqref{eq:increment} gives
$z_d^{1+\gamma}-z_{d-1}^{1+\gamma}\in[(1+\gamma)(z_{d-1}/z_d)^{\gamma},1+\gamma]$. Since the
increments tend to $0$ while $z_d\to\infty$, $z_{d-1}/z_d\to1$, so
$z_d^{1+\gamma}=(1+\gamma)(d+1)(1+o(1))$ and
$z_d-z_{d-1}=(z_d^{1+\gamma})^{-\rho}\sim(1+\gamma)^{-\rho}d^{-\rho}$.
\end{proof}

\neardiagonal*
\begin{proof}
(1) The increments of column $j=t-d$ in rows $j,\dots,t$ are determined by
$\lambda_j,\dots,\lambda_t$ through~\eqref{eq:row}, continuously. If all equal $\lambda$,
homogeneity ($\lambda\mapsto r\lambda$ maps $x\mapsto r^{1-\rho}x$) and
Lemma~\ref{lem:zd} give $P_{t,t-d}=\lambda^{1-\rho}(z_d-z_{d-1})$. By
Theorem~\ref{thm:lambda}, $\lambda_i/\lambda_t\to1$ uniformly for $i\in[t-d,t]$ and
$\lambda_t^{1-\rho}=(\astar/t)^{1-\rho}(1+o(1))$; continuity gives the claim.

(2) By Lemma~\ref{lem:ratio}(2), $\eta_{tj}\to0$ as $d\to\infty$, uniformly in $t$. By
Theorem~\ref{thm:lambda}, uniformly for $i\in[t-d,t]$, $\lambda_i=(\astar+o(1))/i$, so
$\Lam_t-\Lam_{j-1}=\astar(d+1)t^{-1}(1+o(1))$ when $d/t\to0$. Then
\[
  P_{tj}=\lambda_t\bigl((1+\gamma)(\Lam_t-\Lam_{j-1})(1-\eta_{tj})\bigr)^{-\rho}
  =t^{\rho-1}(d+1)^{-\rho}\astar^{\,1-\rho}(1+\gamma)^{-\rho}(1+o(1)),
\]
and $\astar^{\,1-\rho}(1+\gamma)^{-\rho}=1/\Gamma(1-\rho)$. The two regimes match by
Lemma~\ref{lem:zd}.
\end{proof}

\begin{remark}
\label{rem:window}
Let $N_t(L)=\sum_{t-L<j\le t}P_{tj}$ be the mass within distance $L$ of the diagonal, with
constant scores.
\emph{Fixed window.} By Theorem~\ref{thm:near}(1) for $d=0,\dots,L-1$ the increments
telescope, $t^{1-\rho}N_t(L)\to\astar^{\,1-\rho}\sum_{d<L}(z_d-z_{d-1})=\astar^{\,1-\rho}z_{L-1}$.
By Lemma~\ref{lem:zd}, $z_{L-1}=((1+\gamma)L)^{1-\rho}(1+o(1))$ as $L\to\infty$, and
$\astar^{\,1-\rho}(1+\gamma)^{1-\rho}=(1+\gamma)/\Gamma(1-\rho)=1/\Gamma(2-\rho)$, so the limit
is $L^{1-\rho}/\Gamma(2-\rho)\,(1+o(1))$. Thus the rate $t^{-(1-\rho)}$ of
Theorem~\ref{thm:dilution} is attained, while the dependence on $L$ is $L^{1-\rho}$ rather
than the bound's $L$.
\emph{Proportional window.} For $L=\sigma t$, $N_t(L)\to\int_{1-\sigma}^{1}f$: on
$[1-\sigma,1-\vartheta]$ by uniform convergence (Theorem~\ref{thm:shape}), and the range
$j>(1-\vartheta)t$ contributes $O(\vartheta^{1-\rho})$ by~\eqref{eq:const-profile},
Lemma~\ref{lem:row-ends} and $\lambda_t\le1/t$. As $\sigma\to0$,
$\int_{1-\sigma}^{1}f=\sigma^{1-\rho}/\Gamma(2-\rho)\,(1+o(1))$ since
$f(s)\sim(1-s)^{-\rho}/\Gamma(1-\rho)$, which matches the fixed window with $L=\sigma t$.
\end{remark}

\subsection{Proof of Theorem~\ref{thm:iters} (finite iterations)}
\label{app:ot-iters}

\finiteiterations*

The proof is an induction on $k$, in which each step passes from the scaling limit of
the scalings $v^{(k)}$ to that of the row, and from the row to the column mass. The two
passages do not depend on $k$, and we state them first as lemmas about arrays.
Write $\ell(s)=\log(1/s)$, so that $f=\ell^{-\rho}/\Gamma(1-\rho)$, and
\[
  G(s)=\int_s^1f(u)\,\frac{du}u=\frac1{\Gamma(1-\rho)}\int_0^{\ell(s)}x^{-\rho}\,dx
  =\frac{\ell(s)^{1-\rho}}{(1-\rho)\Gamma(1-\rho)}\qquad(x=\ell(u),\ du/u=-dx).
\]
``Uniformly'' means uniformly in $s\in[\delta,1-\delta]$ as $t\to\infty$, with
$j=\lfloor st\rfloor$, for every $\delta\in(0,\frac12)$; every compact subset of $(0,1)$
lies in such an interval, and on it $\ell_{t,\lfloor st\rfloor}\to \ell(s)$ uniformly and
$\ell(s)\ge\log\frac1{1-\delta}>0$.

\begin{lemma}
\label{lem:row-limit}
Let $v_{tj}\ge0$ with $v_{tj}\le C\ell_{tj}^{-\rho}$ for all $j\le t$, and
$v_{t,\lfloor st\rfloor}\to V\,\ell(s)^{-\rho}$ uniformly, for a constant $V\ge0$. Then
$\Sigma_t=t^{-1}\sum_{j\le t}v_{tj}\to V\Gamma(1-\rho)$.
\end{lemma}
\begin{proof}
By (I), $V\Gamma(1-\rho)=V\int_0^1\ell^{-\rho}$. Fix $\vartheta\in(0,\frac12)$ and split both
the sum and the integral at $\vartheta$ and $1-\vartheta$:
$\Sigma_t=A_{\rm far}+A_{\rm mid}+A_{\rm near}$ over $j<\vartheta t$,
$j\in J_t(\vartheta)$, $j>(1-\vartheta)t$, and
$\int_0^1\ell^{-\rho}=I_{\rm far}+I_{\rm mid}+I_{\rm near}$ over $[0,\vartheta]$,
$[\vartheta,1-\vartheta]$, $[1-\vartheta,1]$. With the Riemann sum
$R_t=t^{-1}\sum_{j\in J_t(\vartheta)}\ell(j/t)^{-\rho}$, the triangle inequality gives
\[
  \bigl|\Sigma_t-V\Gamma(1-\rho)\bigr|\le
  \underbrace{|A_{\rm mid}-VR_t|}_{(1)}+\underbrace{V|R_t-I_{\rm mid}|}_{(2)}
  +\underbrace{A_{\rm far}+A_{\rm near}}_{(3)}
  +\underbrace{V(I_{\rm far}+I_{\rm near})}_{(4)} .
\]
Only (1) uses the convergence: it is at most
$\max_{j\in J_t(\vartheta)}|v_{tj}-V\ell(j/t)^{-\rho}|\to0$. (2) tends to zero because
$\ell^{-\rho}$ is continuous on $[\vartheta,1-\vartheta]$. (3) is at most
$C\vartheta(\log\frac1\vartheta)^{-\rho}+2C\vartheta^{1-\rho}/(1-\rho)$ by
Lemma~\ref{lem:row-ends}. (4) is elementary: $\ell\ge\log\frac1\vartheta$ on $[0,\vartheta]$
and $\ell(s)\ge1-s$ give $I_{\rm far}\le\vartheta(\log\frac1\vartheta)^{-\rho}$ and
$I_{\rm near}\le\vartheta^{1-\rho}/(1-\rho)$. Hence
\[
  \limsup_{t\to\infty}\bigl|\Sigma_t-V\Gamma(1-\rho)\bigr|
  \le(V+C)\Bigl(\vartheta\bigl(\log\tfrac1\vartheta\bigr)^{-\rho}
  +\frac{2\vartheta^{1-\rho}}{1-\rho}\Bigr)
\]
for every $\vartheta\in(0,\frac12)$; the left side does not depend on $\vartheta$.
\end{proof}

\begin{lemma}
\label{lem:col-limit}
Let $P_{ij}\ge0$ for $j\le i$, with $iP_{ij}\le C\ell_{ij}^{-\rho}$ for all $j\le i$ and
$tP_{t,\lfloor st\rfloor}\to f(s)$ uniformly. Then
$M_{t,\lfloor st\rfloor}=\sum_{i=\lfloor st\rfloor}^{t}P_{i,\lfloor st\rfloor}\to G(s)$
uniformly.
\end{lemma}
\begin{proof}
Fix $\delta$ and let $s\in[\delta,1-\delta]$, $j=\lfloor st\rfloor$. Column $j$ is seen by
rows $i=j,\dots,t$, and row $i$ sees it at relative position $u_i=j/i\in[j/t,1]$:
$M_{tj}=\sum_{i=j}^{t}i^{-1}(iP_{ij})$. Fix $\vartheta\in(0,\frac12)$ and split the rows at
$u_i=1-\vartheta$.

\emph{Rows with $u_i\le1-\vartheta$.} Here $u_i\ge j/t\ge s-1/t\ge\delta/2$ for
$t\ge2/\delta$, so $u_i$ lies in the fixed interval $[\delta/2,1-\vartheta]$, and every row
index $i\ge j\ge\delta t-1$ tends to infinity. The hypothesis, applied with
$\min(\delta/2,\vartheta)$ in place of $\delta$, gives $iP_{ij}=f(j/i)+o(1)$ uniformly in
$i$ and $s$. The error contributes $o(1)\sum_{i=j}^{t}1/i=o(1)$, since this sum is at most
$\log\frac1\delta+1$. The main part is $\sum_ig(i)$ with $g(x)=f(j/x)/x$, which is positive
and decreasing in $x$ because $f$ is increasing. By (M) it differs from
\[
  \int_{j/(1-\vartheta)}^{t}\frac{f(j/x)}{x}\,dx=\int_{j/t}^{1-\vartheta}f(u)\,\frac{du}u
  \qquad(u=j/x)
\]
by at most two terms of size $g(j/(1-\vartheta))\le f(1-\vartheta)/j\to0$ (one from (M),
one from rounding the lower limit to an integer), and the right side tends to
$\int_s^{1-\vartheta}f(u)\,du/u$ uniformly because $j/t\to s$ uniformly.

\emph{Rows with $u_i>1-\vartheta$}, i.e.\ $j\le i<j/(1-\vartheta)$, are the rows nearest
the diagonal of column $j$; here the bound on $iP_{ij}$ replaces the convergence. With
$r=i+1-j$, $\ell_{ij}\ge1-\frac j{i+1}=\frac r{i+1}$ gives
$P_{ij}\le\frac Ci\bigl(\frac{i+1}{r}\bigr)^{\rho}\le\frac Cj(3j)^{\rho}r^{-\rho}$, since
$j\le i$ and $i+1\le2j+1\le3j$. The range is
$r<\frac{\vartheta j}{1-\vartheta}+1\le2\vartheta j+1\le3\vartheta j$ once $j\ge1/\vartheta$,
and (M) for the decreasing $x\mapsto x^{-\rho}$ gives
$\sum_{r=1}^{R}r^{-\rho}\le R^{1-\rho}/(1-\rho)$, so
\[
  \sum_{j\le i<j/(1-\vartheta)}P_{ij}
  \le\frac Cj(3j)^{\rho}\,\frac{(3\vartheta j)^{1-\rho}}{1-\rho}
  =\frac{3C\vartheta^{1-\rho}}{1-\rho},
\]
in which the powers of $j$ cancel. The missing integral $\int_{1-\vartheta}^{1}f(u)\,du/u$
is at most $2\vartheta^{1-\rho}/((1-\rho)\Gamma(1-\rho))$, by $u\ge\frac12$ and
$\ell(u)\ge1-u$. Letting $\vartheta\to0$ gives $M_{tj}\to G(s)$ uniformly.
\end{proof}

\emph{Strategy.} The proof does not use the fixed point. With constant scores every
iterate is an explicit function of the previous one, and an induction on $k$ shows first
that every iterate has the order of the limit shape (Step~1) and then, with
Lemmas~\ref{lem:row-limit} and~\ref{lem:col-limit}, that it converges to it (Step~2).

Unrolling~\eqref{eq:iteration} with $v^{(1)}=1$ gives
$\log v^{(k)}=-\sum_{m=1}^{k-1}\rho^{k-m}\log M^{(m)}$; with $K\equiv1$,
$P^{(k)}=\rownorm(v^{(k)})$. Write $h_{tj}=\sum_{i=j}^{t}1/i$ for the harmonic sums. For
$\theta\in[0,1)$ we use:
\begin{itemize}
  \item[(R0)] $\ell_{tj}\le h_{tj}\le3\ell_{tj}$ (integral comparison; and
  $h_{tj}\le1/j+\log(t/j)$ with $\ell_{tj}\ge1/(2j)$);
  \item[(R1)] $\sum_{j=1}^{t}\ell_{tj}^{-\theta}\asymp t$, and it equals
  $(t+1)\Gamma(1-\theta)(1+o(1))$ by (M) and (I);
  \item[(R2)] $\sum_{i=j}^{t}\ell_{ij}^{-\theta}/i\asymp\ell_{tj}^{1-\theta}$ uniformly in
  $1\le j\le t$ (a Riemann sum for $\int u^{-\theta}du$; the first term, $i=j$, has the
  order of $\ell_{jj}^{1-\theta}$).
\end{itemize}
Here $\asymp$ means two-sided bounds with constants depending only on $\theta$ (and below
on $k,\rho$).

\emph{Step 1: two-sided bounds.} We show by induction on $k\ge2$ that
$v^{(k)}\asymp\ell^{-\rho}$, $P^{(k)}\asymp\ell^{-\rho}/t$ and $M^{(k)}\asymp\ell^{1-\rho}$.
For $k=1$, $P^{(1)}_{tj}=1/t$ and $M^{(1)}_{tj}=h_{tj}\asymp\ell_{tj}$ by (R0), so
$v^{(2)}=(M^{(1)})^{-\rho}\asymp\ell^{-\rho}$. From $v^{(k)}\asymp\ell^{-\rho}$, (R1) gives
$P^{(k)}\asymp\ell^{-\rho}/t$ and (R2) gives $M^{(k)}\asymp\ell^{1-\rho}$; then
$v^{(k+1)}=(v^{(k)}/M^{(k)})^{\rho}\asymp(\ell^{-\rho}\ell^{-(1-\rho)})^{\rho}=\ell^{-\rho}$.

\emph{Step 2: limits.} We show by induction on $k\ge2$ that for a constant $V_k>0$
\[
  (\mathcal I_k)\qquad
  v^{(k)}_{t,\lfloor st\rfloor}\to V_k\ell(s)^{-\rho}\quad\text{uniformly},
\]
and, in the induction step, that $tP^{(k)}_{t,\lfloor st\rfloor}\to f(s)$ uniformly,
which is the theorem. Step~1 provides a constant $C$ with $v^{(k)}_{tj}\le C\ell_{tj}^{-\rho}$
and $tP^{(k)}_{tj}\le C\ell_{tj}^{-\rho}$ for all $j\le t$; these bounds control the ends
of a row, where the induction hypothesis says nothing.

\emph{Base case $k=2$.} $v^{(1)}=1$, so $P^{(1)}_{tj}=1/t$ and
$M^{(1)}_{tj}=h_{tj}$. By (H), $\log\frac{t+1}{j}\le h_{tj}\le\frac1j+\log\frac tj$,
and $j\ge\delta t-1$, so $M^{(1)}_{t,\lfloor st\rfloor}\to \ell(s)$ uniformly. Since $\ell$ is bounded
below by a positive constant and $x\mapsto x^{-\rho}$ is uniformly continuous away from
zero, $v^{(2)}=(v^{(1)}/M^{(1)})^{\rho}=(M^{(1)})^{-\rho}$ satisfies $(\mathcal I_2)$ with
$V_2=1$.

\emph{Induction step.} Assume $(\mathcal I_k)$.
\begin{enumerate}
  \item[(a)] \emph{Row normalizer and shape.} By Step~1, $(\mathcal I_k)$ and
  Lemma~\ref{lem:row-limit}, $\Sigma_t=t^{-1}\sum_{j\le t}v^{(k)}_{tj}\to V_k\Gamma(1-\rho)>0$.
  Since $P^{(k)}_{tj}=v^{(k)}_{tj}/(t\Sigma_t)$,
  \[
    tP^{(k)}_{t,\lfloor st\rfloor}=\frac{v^{(k)}_{t,\lfloor st\rfloor}}{\Sigma_t}
    \to\frac{V_k\ell(s)^{-\rho}}{V_k\Gamma(1-\rho)}=f(s)\quad\text{uniformly}.
  \]
  \item[(b)] \emph{Column mass.} By Step~1, (a) and Lemma~\ref{lem:col-limit},
  $M^{(k)}_{t,\lfloor st\rfloor}\to G(s)=\ell(s)^{1-\rho}/((1-\rho)\Gamma(1-\rho))$ uniformly.
  \item[(c)] \emph{Next iterate.} On $[\delta,1-\delta]$, $G\ge G(1-\delta)>0$, so the
  quotient
  $v^{(k)}/M^{(k)}$ converges uniformly to $V_k\ell^{-\rho}/G$, and $x\mapsto x^{\rho}$ is
  uniformly continuous on compact subsets of $(0,\infty)$. Hence
  \[
    v^{(k+1)}_{t,\lfloor st\rfloor}=\Bigl(\frac{v^{(k)}}{M^{(k)}}\Bigr)^{\rho}_{t,\lfloor st\rfloor}
    \to\Bigl(V_k(1-\rho)\Gamma(1-\rho)\Bigr)^{\rho}\ell(s)^{-\rho\cdot\rho-\rho(1-\rho)}
    =V_{k+1}\ell(s)^{-\rho},
  \]
  with $V_{k+1}=(V_k(1-\rho)\Gamma(1-\rho))^{\rho}>0$, which is $(\mathcal I_{k+1})$.
\end{enumerate}
The exponent identity $\rho\cdot\rho+\rho(1-\rho)=\rho$ keeps the shape fixed from $k=2$ on;
only $V_k$ changes, and it cancels in the row normalization of (a). \hfill$\square$

\subsection{Proof of Proposition~\ref{prop:iid} (i.i.d.\ content)}
\label{app:iid}

\iidcontent*

Let $K=e^{S/\eps}$, with i.i.d.\ entries in $[e^{-B},e^{B}]$, and $\kappa=\E K$. Write
\[
  W_{tj}=\sum_{i=j}^{t}\lambda_iK_{ij},\qquad \lambda'_t=\kappa\lambda_t,\qquad
  \Lam'_t=\kappa\Lam_t ,
\]
so that Lemma~\ref{lem:sandwich} reads $W_{tj}\le M_{tj}^{1+\gamma}\le(1+\gamma)W_{tj}$.
Constants $C_B$ below depend only on $B$ and $\rho$ and may change from line to line;
$\mathcal F_t$ is the $\sigma$-field generated by rows $1,\dots,t$ of $K$.

\emph{Strategy.} In the constant-score proof, $\Lam_t-\Lam_{j-1}$ plays two roles: it is
the column sum that controls $M_{tj}$, and it is built from the row normalizers. With
content, the column sum is $W_{tj}$ and the row normalization weights each column by
$K_{tj}$. Both sums average independent entries: $W_{tj}=\Lam'_t-\Lam'_{j-1}+o(1)$
(Lemma~\ref{lem:col-lln}) and $\sum_jK_{tj}M_{tj}^{-\gamma}=\kappa\sum_jM_{tj}^{-\gamma}+o(t)$
(Lemma~\ref{lem:row-lln}), almost surely and uniformly over the relevant ranges. The
rescaled normalizers $\lambda'$ then satisfy the constant-score equations up to vanishing
errors, which are the hypotheses of the bootstrap Lemma~\ref{lem:bootstrap}. The column
law of large numbers needs two tools: a deterministic bound on how much one entry moves a
row's normalizer (Lemma~\ref{lem:loo}), and a maximal concentration inequality for
martingales (Lemma~\ref{lem:azuma}).

\emph{Deterministic bounds.} Every $K$ with entries in $[e^{-B},e^{B}]$ satisfies
Lemmas~\ref{lem:wellposed}--\ref{lem:lambda-upper}, with $\lambda_t\le e^{B}/t$, and the
bounded-score Lemmas~\ref{lem:alpha0} and~\ref{lem:ratio}:
\begin{enumerate}
  \item[(D1)] $t\lambda_t\ge a_B$ (Lemma~\ref{lem:alpha0}).
  \item[(D2)] $r_i\le c_B/(i-j+1)$, and for every $\vartheta\in(0,\frac12)$,
  $\bar\eta_t=\max_{\vartheta t\le j\le(1-\vartheta)t}\eta_{tj}\to0$ (Lemma~\ref{lem:ratio}(1)).
  \item[(D3)] For $j\le(1-\vartheta)t$, $W_{tj}\ge e^{-B}a_B\log\frac{t+1}{j}
  \ge e^{-B}a_B\vartheta$; for every $j\ge\vartheta t$, $W_{tj}\le e^{2B}(1+\log\frac1\vartheta)$,
  hence $M_{tj}\le C_B(1+\log\frac1\vartheta)$.
  \item[(D4)] The near and far parts of a row are small: by Lemma~\ref{lem:sandwich},
  $K_{tj}M_{tj}^{-\gamma}\le e^{B}W_{tj}^{-\rho}\le C_B\ell_{tj}^{-\rho}$, since
  $W_{tj}\ge e^{-B}a_B\log\frac{t+1}j$ by (D1) and (H); so by Lemma~\ref{lem:row-ends},
  for $t\ge1/\vartheta$ the parts with $j<\vartheta t$ and $j>(1-\vartheta)t$ sum to at most
  $C_Bt\,\chi(\vartheta)$.
\end{enumerate}

\emph{Two tools.} Write $\Phi(C,k)$ for the normalizer of a row with masses
$C=(C_{j'})_{j'\le i}$ and entries $k=(k_{j'})_{j'\le i}$, that is, the unique $\lambda$
with $F(\lambda;C,k)=1$, where $F(\lambda;C,k)=\sum_{j'}x_{j'}$ and
$x_{j'}(C_{j'}+x_{j'})^{\gamma}=\lambda k_{j'}$ (Lemma~\ref{lem:wellposed}).

\begin{lemma}
\label{lem:loo}
Let $C\in[0,\infty)^i$, and let $k,k'\in[e^{-B},e^{B}]^i$ differ only in the coordinate $j$.
If $x=(x_{j'})$ solves the row for $(C,k)$ at $\lambda=\Phi(C,k)$, then
\[
  \bigl|\log\Phi(C,k')-\log\Phi(C,k)\bigr|\le(1+\gamma)e^{2B}x_j .
\]
\end{lemma}
\begin{proof}
The \emph{elasticity} $d\log x/d\log\lambda$, the relative change of $x$ per relative
change of $\lambda$, follows by differentiating $\log x+\gamma\log(C+x)=\log\lambda+\log k$:
\[
  \frac{d\log x}{d\log\lambda}=\frac{d\log x}{d\log k}=\frac{C+x}{C+(1+\gamma)x}
  \in\Bigl[\frac1{1+\gamma},\,1\Bigr],
\]
the same in $k$ because $\lambda$ and $k$ enter only through $\lambda k$. It is used twice.
First, at $\lambda=\Phi(C,k)$, replacing $k_j$ by $k'_j$ leaves every $x_{j'}$ with
$j'\neq j$ unchanged and multiplies $x_j$ by a factor $m$ with
$|\log m|\le|\log(k'_j/k_j)|\le2B$, since the elasticity lies in $(0,1]$; so the row sum
becomes
\[
  F=1+(m-1)x_j\in\bigl[1-(1-e^{-2B})x_j,\ 1+(e^{2B}-1)x_j\bigr],
\]
and $|\log F|\le e^{2B}x_j$: above one by $\log(1+u)\le u$, below one by
$-\log(1-u)\le u/(1-u)$ with $u=(1-e^{-2B})x_j$ and $1-u\ge e^{-2B}$. Second, the
elasticity of the row sum, $d\log F/d\log\lambda=\sum_{j'}(x_{j'}/F)\,d\log x_{j'}/d\log\lambda$,
is a weighted mean of elasticities and so at least $1/(1+\gamma)$; restoring the row sum
to one therefore moves $\log\lambda$ by at most $(1+\gamma)|\log F|$.
\end{proof}

\begin{lemma}
\label{lem:azuma}
Let $S_t=\sum_{i=t_0}^{t}X_i$, $t\ge t_0$, be a martingale with respect to a filtration
$(\mathcal F_t)$, with $|X_i|\le c_i$ and $V=\sum_{i\ge t_0}c_i^2<\infty$. Then for $u>0$,
$\Pr(\sup_{t\ge t_0}|S_t|\ge u)\le2\exp(-u^2/(2V))$.
\end{lemma}
This is the maximal form of the Azuma--Hoeffding
inequality~\citep{hoeffding1963probability,azuma1967weighted}: Hoeffding's lemma gives
$\E e^{\theta S_t}\le e^{\theta^2V/2}$, and Doob's maximal inequality for the submartingale
$e^{\theta S_t}$ with $\theta=u/V$, applied to $S$ and $-S$, gives the bound.

\begin{lemma}
\label{lem:col-lln}
Almost surely,
\begin{equation}
  \max_{\vartheta t\le j\le t}\bigl|W_{tj}-(\Lam'_t-\Lam'_{j-1})\bigr|\to0
  \qquad\text{for every }\vartheta\in(0,1).
  \label{eq:col-lln}
\end{equation}
\end{lemma}
\begin{proof}
The difference is $\sum_{i=j}^{t}\lambda_i(K_{ij}-\kappa)$. Its weights depend on the
summands: $\lambda_i$ is computed from row $i$, which contains $K_{ij}$. We remove this
dependence by leaving one entry out. For $i\ge j$ let
\[
  \lambda_i^{(-j)}=\Phi\bigl(M_{i-1,\cdot},\,(K_{i1},\dots,K_{i,j-1},\kappa,K_{i,j+1},\dots,K_{ii})\bigr),
\]
the normalizer of row $i$ with the actual masses and entries except that $K_{ij}$ is
replaced by $\kappa$. Let $\mathcal G_i=\sigma(\mathcal F_{i-1},K_{ij'}:j'\neq j)$, so that
$\mathcal F_{i-1}\subset\mathcal G_i\subset\mathcal F_i$ and, the entries being independent,
$K_{ij}$ is independent of $\mathcal G_i$. Then $\lambda_i^{(-j)}$ is
$\mathcal G_i$-measurable: by the row recursion, $M_{i-1,\cdot}$ is a measurable function of
rows $1,\dots,i-1$ of $K$; and $\Phi$ is Borel, because $F$ is continuous and increasing in
$\lambda$ (proof of Lemma~\ref{lem:wellposed}), so $\{\Phi\le c\}=\{F(c;\cdot)\ge1\}$ is
closed. The modified row still has unit mass on columns $1,\dots,i$, so the proof of
Lemma~\ref{lem:lambda-upper} gives $\lambda_i^{(-j)}\le e^{B}/i$. Define
\[
  S^{(j)}_t=\sum_{i=j}^{t}X_i,\quad X_i=\lambda_i^{(-j)}(K_{ij}-\kappa),\qquad
  R^{(j)}_t=\sum_{i=j}^{t}\bigl(\lambda_i-\lambda_i^{(-j)}\bigr)(K_{ij}-\kappa),
\]
so that $W_{tj}-(\Lam'_t-\Lam'_{j-1})=S^{(j)}_t+R^{(j)}_t$: a martingale part, whose
weights are decoupled from the summands, and a leave-one-out error.

\emph{The martingale part.} Since $\lambda_i^{(-j)}$ is $\mathcal G_i$-measurable and
bounded, and $K_{ij}$ is independent of $\mathcal G_i$ with mean $\kappa$,
\[
  \E[X_i\mid\mathcal G_i]=\lambda_i^{(-j)}\,\E[K_{ij}-\kappa]=0,
  \qquad\text{hence}\qquad\E[X_i\mid\mathcal F_{i-1}]=0
\]
by the tower property. Moreover $X_i$ is $\mathcal F_i$-measurable, and since $\kappa$ and
$K_{ij}$ lie in $[e^{-B},e^{B}]$,
\[
  |X_i|\le\frac{e^{B}}{i}\,(e^{B}-e^{-B})\le c_i=\frac{e^{2B}}{i},\qquad
  V_j=\sum_{i\ge j}c_i^2\le\frac{e^{4B}}{j-1}\quad(j\ge2).
\]
So $(S^{(j)}_t)_{t\ge j}$ is a martingale with respect to $(\mathcal F_t)$, and by
Lemma~\ref{lem:azuma},
\[
  \Pr\Bigl(\sup_{t\ge j}\bigl|S^{(j)}_t\bigr|\ge u\Bigr)
  \le2\exp\Bigl(-\frac{u^2}{2V_j}\Bigr)\le2\exp\Bigl(-\frac{u^2(j-1)}{2e^{4B}}\Bigr)
  \qquad(u>0).
\]
The right side is summable over $j$. By the Borel--Cantelli lemma, almost surely the
event on the left occurs for only finitely many $j$; applying this to $u=1/n$ for
every $n\ge1$ gives, almost surely,
\[
  \sup_{j\ge J}\ \sup_{t\ge j}\bigl|S^{(j)}_t\bigr|\to0\qquad(J\to\infty).
\]

\emph{The leave-one-out error.} Row $i$ solves the row equation with masses
$C=M_{i-1,\cdot}$, entries $K_{i\cdot}$ and solution $x_{j'}=P_{ij'}$, and
$\lambda_i^{(-j)}$ changes only the entry $j$, from $K_{ij}$ to $\kappa\in[e^{-B},e^{B}]$.
By Lemma~\ref{lem:loo},
$|\log\lambda_i^{(-j)}-\log\lambda_i|\le(1+\gamma)e^{2B}P_{ij}$, hence
$|\lambda_i-\lambda_i^{(-j)}|\le C_B\lambda_iP_{ij}$, and with $\lambda_i\le e^{B}/i\le
e^{B}/j$,
\[
  \bigl|R^{(j)}_t\bigr|\le C_B\sum_{i=j}^{t}\lambda_iP_{ij}\le\frac{C_B}{j}\sum_{i=j}^{t}P_{ij}
  =\frac{C_B\,M_{tj}}{j}\le\frac{C_B(1+\log\frac1\vartheta)}{\vartheta t}
\]
for $j\ge\vartheta t$, by (D3).
\end{proof}

\begin{lemma}
\label{lem:row-lln}
Almost surely, for every $\vartheta\in(0,\frac12)$,
\begin{equation}
  \frac1t\sum_{\vartheta t\le j\le(1-\vartheta)t}(K_{tj}-\kappa)\,M_{t-1,j}^{-\gamma}\to0 ;
  \label{eq:row-lln}
\end{equation}
and for all $K$ and $\vartheta t\le j\le(1-\vartheta)t$,
\begin{equation}
  \Bigl(1-\frac{\max(1,\gamma)c_B}{\vartheta t}\Bigr)M_{t-1,j}^{-\gamma}\le M_{tj}^{-\gamma}
  \le M_{t-1,j}^{-\gamma}.
  \label{eq:own-term}
\end{equation}
\end{lemma}
\begin{proof}
The weights are $M_{t-1,j}^{-\gamma}$ rather than the $M_{tj}^{-\gamma}$ of the row
normalization: $M_{tj}$ contains the own term $P_{tj}$, which depends on $K_{tj}$, whereas
$M_{t-1,j}$ is $\mathcal F_{t-1}$-measurable and the entries $K_{tj}$ of row $t$ are
independent of $\mathcal F_{t-1}$. In the middle range the weights are bounded: by
Lemma~\ref{lem:sandwich} and (D3) applied to row $t-1$,
$M_{t-1,j}^{-\gamma}\le W_{t-1,j}^{-\rho}\le(e^{B}/(a_B\vartheta))^{\rho}=c_w$; near the
diagonal they are not, and at $j=t$ the mass $M_{t-1,t}$ is zero. Conditionally on
$\mathcal F_{t-1}$ the sum is a sum of at most $t$ independent centered terms, each in an
interval of length $2e^{B}c_w$, so Hoeffding's inequality~\citep{hoeffding1963probability}
bounds the probability that it exceeds $ut$ by $2\exp(-u^2t/(2e^{2B}c_w^2))$ for $u>0$,
which is summable in $t$; Borel--Cantelli over $t$ and over rational $u,\vartheta$
gives~\eqref{eq:row-lln}. The own term of row $t$ changes the weights only by a factor:
$M_{tj}=M_{t-1,j}/(1-r_t)$ with $r_t\le c_B/(t-j+1)\le c_B/(\vartheta t)$ by (D2), and
$(1-r)^{\gamma}\ge1-\max(1,\gamma)r$ gives~\eqref{eq:own-term}.
\end{proof}

\emph{Proof of Proposition~\ref{prop:iid}.} \emph{The normalizers.} Fix an outcome in the
almost-sure events of Lemmas~\ref{lem:col-lln} and~\ref{lem:row-lln}; everything below is
deterministic. We apply Lemma~\ref{lem:bootstrap} to
$\lambda'_t=\kappa\lambda_t$, so that $D_{tj}=\Lam'_t-\Lam'_{j-1}$ and
$1/\lambda'_t=\kappa^{-1}\sum_jK_{tj}M_{tj}^{-\gamma}$. Compared with
Theorem~\ref{thm:lambda} there are three errors, each controlled by an earlier step: the
content weights $K_{tj}$ in the row sum~\eqref{eq:row-lln}, the own term of row $t$
\eqref{eq:own-term}, and the gap between $W$ and $D$~\eqref{eq:col-lln}.

By Lemmas~\ref{lem:lambda-upper} and~\ref{lem:alpha0}, $\kappa a_B\le t\lambda'_t\le\kappa e^{B}$.
The lemma asks for (i) and (ii) for every $\vartheta\in(0,\frac12)$ and every tolerance
$\nu_0\in(0,\frac12)$, so fix both. For $j\in J_t(\vartheta)$, Lemma~\ref{lem:alpha0} and (H)
give $D_{tj}\ge\kappa a_B\log\frac{t+1}j\ge\kappa a_B\vartheta=d_\vartheta$. Each of the
three errors above tends to zero, and we make them small relative to a tolerance
\[
  \nu=\frac{\nu_0}{c_\vartheta}\le\frac{\nu_0}2,\qquad
  c_\vartheta=\max\Bigl(2,\ \frac1\kappa+\bigl((1+\gamma)d_\vartheta/2\bigr)^{-\rho}\Bigr),
\]
chosen so that the errors add up to at most $\nu_0$ in (i) and (ii). Since $D_{tj}\ge d_\vartheta$,
an additive error of at most $\nu d_\vartheta$ in $D_{tj}$ is a factor in $[1-\nu,1+\nu]$,
and, as $\nu<\frac12$,
\begin{equation}
  \sum_{j\in J_t(\vartheta)}\bigl((1+\gamma)(1\pm\nu)D_{tj}\bigr)^{-\rho}
  \le t\bigl((1+\gamma)d_\vartheta/2\bigr)^{-\rho}.
  \label{eq:mid-size}
\end{equation}
For large $t$, depending on $\vartheta$, $\nu$ and the fixed outcome, the following hold
for all $j\in J_t(\vartheta)$: by~\eqref{eq:col-lln} at rows
$t$ and $t-1$ and $D_{tj}-D_{t-1,j}=\lambda'_t\le\kappa e^{B}/t$, both $W_{tj}$ and
$W_{t-1,j}$ differ from $D_{tj}$ by at most $\nu d_\vartheta$; by
Lemma~\ref{lem:ratio}(1) at row $t-1$ with $\vartheta/2$ (note
$J_t(\vartheta)\subset J_{t-1}(\vartheta/2)$), $\bar\eta_{t-1}\le\nu$; the factor
in~\eqref{eq:own-term} is at least $1-\nu$; and the sum in~\eqref{eq:row-lln} is at most
$\nu t$ in absolute value.

\emph{Hypothesis (i).} Keeping the middle range and using~\eqref{eq:own-term},
\eqref{eq:row-lln}, $M_{t-1,j}^{-\gamma}\ge M_{tj}^{-\gamma}\ge((1+\gamma)W_{tj})^{-\rho}$
(Lemma~\ref{lem:sandwich}) and $W_{tj}\le(1+\nu)D_{tj}$,
\[
  \frac1{\lambda'_t}\ge\frac{1-\nu}{\kappa}\sum_{j\in J_t(\vartheta)}K_{tj}M_{t-1,j}^{-\gamma}
  \ge(1-\nu)\sum_{j\in J_t(\vartheta)}\bigl((1+\gamma)(1+\nu)D_{tj}\bigr)^{-\rho}
  -\frac{\nu t}\kappa .
\]
By~\eqref{eq:mid-size}, the factor $1-\nu$ costs at most $\nu t((1+\gamma)d_\vartheta/2)^{-\rho}$,
so the total additive error is at most
$\nu t\bigl(\frac1\kappa+((1+\gamma)d_\vartheta/2)^{-\rho}\bigr)\le\nu c_\vartheta t=\nu_0t$;
with $1+\nu\le1+\nu_0$ this is (i) for $\nu_0$.

\emph{Hypothesis (ii).} The near and far parts of $\sum_jK_{tj}M_{tj}^{-\gamma}$ are at most
$C_Bt\chi(\vartheta)$ by (D4). In the middle range, \eqref{eq:own-term}
and~\eqref{eq:row-lln} give
$\sum_{J_t(\vartheta)}K_{tj}M_{tj}^{-\gamma}\le\kappa\sum_{J_t(\vartheta)}M_{t-1,j}^{-\gamma}+\nu t$,
and Lemma~\ref{lem:ratio}(1) at row $t-1$ gives
\[
  M_{t-1,j}^{1+\gamma}\ge(1+\gamma)(1-\bar\eta_{t-1})W_{t-1,j}
  \ge(1+\gamma)(1-\nu)^2D_{tj}.
\]
Dividing by $\kappa$,
$1/\lambda'_t\le\sum_{J_t(\vartheta)}((1+\gamma)(1-\nu)^2D_{tj})^{-\rho}+t(\nu/\kappa+C\chi(\vartheta))$
with $C=C_B/\kappa$. Since $(1-\nu)^2\ge1-2\nu\ge1-\nu_0$ and $\nu/\kappa\le\nu c_\vartheta=\nu_0$,
this is (ii) for $\nu_0$.

Lemma~\ref{lem:bootstrap} gives $t\lambda'_t\to\astar$, that is, $t\lambda_t\to\astar/\kappa$.

\emph{The shape.} Let $j=\lfloor st\rfloor$ with $s\in[\delta,1-\delta]$. By (D2),
$M_{tj}^{1+\gamma}=(1+\gamma)W_{tj}(1-\eta_{tj})$ with $\eta_{tj}\to0$ uniformly. By
\eqref{eq:col-lln}, $W_{tj}=\Lam'_t-\Lam'_{j-1}+o(1)$, and as in the proof of
Theorem~\ref{thm:shape}, $t\lambda'_t\to\astar$ gives $\Lam'_t-\Lam'_{j-1}=\astar\log(1/s)+o(1)$, both
uniformly in $s$. Since $\astar\log(1/s)\ge\astar\log\frac1{1-\delta}>0$,
\[
  tP_{tj}=t\lambda_t\,K_{tj}\,M_{tj}^{-\gamma}
  =\frac{\astar+o(1)}{\kappa}\,K_{tj}\,\bigl((1+\gamma)\astar\log(1/s)\bigr)^{-\rho}(1+o(1))
  =\frac{K_{tj}}{\kappa}f(s)(1+o(1))
\]
uniformly, and $K_{tj}\le e^{B}$ turns the factor $1+o(1)$ into an additive $o(1)$.
\hfill$\square$

\section{Proofs for Section~\ref{sec:potentials}}
\label{app:potentials}

\paragraph{Structure.}
The proofs are short and independent of one another, except that
Proposition~\ref{prop:arch} collects results proved elsewhere. Two arguments recur:
\begin{itemize}
  \item[(B)] \emph{Bounded content.} If the content part of every score lies in
  $[-B,B]$, each softmax weight changes by a factor in $[e^{-B},e^{B}]$ in the numerator
  and in the normalizer, so every ratio of attention masses changes by a factor in
  $[e^{-2B},e^{2B}]$. This reduces Proposition~\ref{prop:rates} and
  Corollary~\ref{cor:boundary} to constant content.
  \item[(Z)] \emph{The normalizer of a power prior.} With constant content and the prior
  $(1+d)^{-c}$, the mass within distance $L$ of the diagonal at row $t$ is $S_L/Z_t$ with
  $Z_t=\sum_{d<t}(1+d)^{-c}=\sum_{k=1}^{t}k^{-c}$, so the question is how $Z_t$ grows.
\end{itemize}

\subsection{Row shifts and finite-dimensional kernels}

\rowshift*
\begin{proof}
Softmax: the numerator and the denominator of row $t$ are multiplied by $e^{r_t}$. Fixed
point: in~\eqref{eq:fp}, $K_{tj}\mapsto e^{r_t}K_{tj}$ is absorbed by
$\lambda_t\mapsto e^{-r_t}\lambda_t$, and $M$ depends only on $P$; by
Lemma~\ref{lem:wellposed} the solution is the same.
\end{proof}

\finiterep*
\begin{proof}
(i) By the Cayley--Hamilton theorem, $b(d)=u^{\top}A^dw$ satisfies a linear recurrence with
constant coefficients, whose solutions are exponential polynomials in the nonzero
eigenvalues of $A$ for $d\ge m$~\citep{everest2003recurrence}.

(ii) Let $r^*$ be the largest $|\mu_k|$ with $p_k\ne0$ and $m^*$ the largest degree among
those terms. Then $b(d)=d^{m^*}r^{*d}(\tau(d)+o(1))$, where
$\tau(d)=\sum_kc_ke^{i\theta_kd}$ has nonzero coefficients and distinct frequencies, so its
mean square tends to $\sum_k|c_k|^2>0$ and $|\tau(d)|\ge\delta>0$ for infinitely many $d$.
Hence $b\to0$ ($r^*<1$), $b$ is bounded ($r^*=1$, $m^*=0$), or $|b(d)|\ge\frac\delta2d$
along a subsequence; none is compatible with $b(d)/\log d\to-c\ne0$.

(iii) The diagonal term bounds the normalizer below, so
$P_{tj}=e^{S_{tj}/\eps}/\sum_{j'\le t}e^{S_{tj'}/\eps}\le e^{B}/(te^{-B})$. For an
orthogonal $R$, $|\langle R(t)q,R(j)k\rangle|\le|q||k|$ by the Cauchy--Schwarz inequality,
which is bounded under bounded inputs.
\end{proof}

\subsection{Far mass}

\farmass*
\begin{proof}
Lemma~\ref{lem:wellposed} gives
$\lambda_t=P_{tt}^{1+\gamma}/K_{tt}\le K_{tt}^{-1}\le e^B$ and
$P_{tj}^{1+\gamma}\le\lambda_tK_{tj}$. Since $1/(1+\gamma)=1-\rho$,
\[
  P_{t,t-d}\le(\lambda_tK_{t,t-d})^{1-\rho}
  \le e^{2(1-\rho)B}e^{-(1-\rho)m_0d}.
\]
Summing over $d\ge L+1$ gives the bound.
\end{proof}

\archbounds*
\begin{proof}
Under bounded inputs every linear function of the input lies in a fixed interval $[-A,A]$,
with $A$ depending on the function. Bounded far mass in the first item follows from
Lemma~\ref{lem:farmass} once $b_{tj}\le-m_0(t-j)$ with $m_0>0$; each such bound on a path
potential follows from a bound on its increments.
\begin{itemize}
  \item \emph{ALiBi}, including OT on ALiBi, is linear with slope $m_0>0$.
  \item \emph{FoX:} $b_{tj}=\sum_{\ell=j+1}^{t}\log\sigma(w^{\top}x_\ell+w_0)$. Under
  bounded inputs $w^{\top}x_\ell+w_0\le A_F$, so each increment is at most
  $\log\sigma(A_F)=-\log(1+e^{-A_F})<0$, which is $-m_0$.
  \item \emph{GRAPE-AP:} each increment is
  $m_h\log\sigma(\langle p_t,R_\ell\rms(p_\ell)\rangle/d_p)$. Since $\rms(p_\ell)$ has norm
  $\sqrt{d_p}$ and $R_\ell$ is orthogonal, the argument is at most
  $|p_t|/\sqrt{d_p}\le A_P$ by the Cauchy--Schwarz inequality, and each increment is at
  most $-m_h\log(1+e^{-A_P})$.
  \item \emph{GRAPE-A:} $b=-\omega_hd(\alpha_t+\beta_j)$, where
  $\omega_h=\softplus(\cdot)\,m_h>0$ and the gates $\alpha_t,\beta_j$ have the form
  $\softplus(\langle\rms(q),\hat v\rangle/\sqrt D)$. The argument lies in
  $[-1,1]$ because $\rms(q)$ has norm $\sqrt D$ and $\hat v$ is a unit vector, so each gate
  lies in $[\softplus(-1),\softplus(1)]=[0.313,1.313]$ and, whatever the inputs,
  $\alpha_t+\beta_j\ge0.626$, so $b\le-0.62\,\omega_hd$.
  \item \emph{Dilution:} RoPE and GRAPE-M by Proposition~\ref{prop:rep}(iii); the OT
  operator by Theorem~\ref{thm:dilution}.
  \item \emph{Log-distance biases:} by Proposition~\ref{prop:rates}, the mass within a
  fixed window tends to zero for $c\le1$, and for $c>1$ the mass beyond $L$ is at most
  $e^{2B}L^{1-c}/(c-1)$ for every $t$. For KERPLE, $(1+ad)/(1+d)$ lies between $\min(a,1)$
  and $\max(a,1)$, so the weights differ from those of $-p\log(1+d)$ by bounded factors and
  the same conclusions hold.
  \item \emph{Query-dependent exponents:} P6 is the case $v=0$, $S_t-S_j=t-j$, of LC,
  $b_{tj}=-c(x_t)\log(1+S_t-S_j)$ with clock ticks $2\sigma(v^{\top}x_\ell)\in[\kappa,2]$,
  $\kappa=2\sigma(-A)>0$, so $\kappa d\le S_t-S_j\le2d$ and $b_{tt}=0$. If
  $c(x)\ge1+\delta$ for all admissible $x$, the weight at distance $d$ is at most
  $e^{2B}(1+\kappa d)^{-1-\delta}$ times the diagonal one, whose tail sum beyond $L$ tends to
  zero uniformly in $t$. If $c(x)\le1$ for all $x$, every weight at distance $d$ is between
  $e^{-2B}(1+2d)^{-1}$ and $e^{2B}$ times the diagonal one, so the mass within $L$ is at most
  $e^{4B}L/\sum_{d<t}(1+2d)^{-1}\to0$.
  \item \emph{CABLE:} $b_{tj}=g_t(S_t-S_j)$ with $S_t-S_j=-\sum_{\ell=j+1}^{t}\max(a_\ell,0)$
  for the clock's pre-activation $a_\ell$, and $g_t=\softplus(\cdot)\ge g_0>0$. If
  $a_\ell\ge a_0>0$ for all admissible inputs, $b\le-g_0a_0(t-j)$ and
  Lemma~\ref{lem:farmass} applies. If $a_\ell\le0$ always, $S$ is constant and $b\equiv0$;
  CABLE has no RoPE, so the scores are bounded content alone and softmax dilutes as in
  Proposition~\ref{prop:rep}(iii).
  \qedhere
\end{itemize}
\end{proof}

\subsection{Power priors}

\powerrates*
\begin{proof}
By (Z) the window mass is $S_L/Z_t$. Comparing the decreasing $k^{-c}$ with its integral,
as in (M) of Appendix~\ref{app:ot}: for $c<1$, $Z_t=t^{1-c}(1-c)^{-1}(1+o(1))$; for $c=1$,
$Z_t=\log t\,(1+o(1))$; for $c>1$, $Z_t$ increases to $\zeta(c)$. For $c>1$ the far mass beyond $L$
is moreover at most
\[
  \frac1{Z_t}\sum_{d\ge L}(1+d)^{-c}\le\sum_{k\ge L+1}k^{-c}\le\int_L^\infty x^{-c}\,dx
  =\frac{L^{1-c}}{c-1}
\]
for every $t$, since $Z_t\ge1$ (the diagonal term). Bounded content enters through (B).
\end{proof}

\boundarycor*
\begin{proof}
By (Z), $N_t(L)=S_L/Z_t$ with $Z_t=\sum_{k=1}^{t}k^{-1-\delta}$. Let
$\phi_\delta(x)=\int_1^{x}y^{-1-\delta}\,dy=(1-x^{-\delta})/\delta$. Since
$x\mapsto x^{-1-\delta}$ is decreasing, integral comparison gives
\[
  \phi_\delta(t+1)\le Z_t\le1+\phi_\delta(t).
\]
Write $u=\delta\log x\ge0$, so that $\phi_\delta(x)=(1-e^{-u})/\delta$; from
$u-\tfrac12u^2\le1-e^{-u}\le\min(u,1)$, dividing by $\delta$ gives
$\log x\,(1-\tfrac12\delta\log x)\le\phi_\delta(x)\le\min(\log x,1/\delta)$.

(i) $Z_t\le1+\phi_\delta(t)\le1+\log t$ gives the lower bound. If $\delta\log(t+1)\le1$,
then $\phi_\delta(t+1)\ge\tfrac12\log(t+1)$, which gives the upper bound.

(ii) $Z_t\le\zeta(c)$ gives the lower bound. The tail satisfies
$\zeta(c)-Z_t=\sum_{k>t}k^{-c}\le\int_t^\infty x^{-c}\,dx=t^{-\delta}/\delta$, and
$Z_t\ge\phi_\delta(t+1)\ge(1-t^{-\delta})/\delta$, so
$\zeta(c)/Z_t\le1+t^{-\delta}/(1-t^{-\delta})=1/(1-t^{-\delta})$, which is the upper bound.
Bounded content enters through (B).
\end{proof}

\section{Numerical checks of the theory}
\label{app:numerics}

We check finite-length identities and bounds (Appendix~\ref{app:num-exact}), compare
finite lengths with the limits (Appendix~\ref{app:num-asymp}), and measure effects the
theory does not quantify (Appendix~\ref{app:num-obs}). The tests are in
\path{packages/geoattn-core/tests/}. ``Recursion'' solves~\eqref{eq:row} row by row in
float64; ``operator'' runs the iteration~\eqref{eq:iteration} in float32.

\subsection{Finite-length identities and bounds}
\label{app:num-exact}

These checks hold entrywise up to rounding, or fail. In \texttt{test\_unbalanced\_recursion}:
the recursion agrees with the operator after 400 iterations to $10^{-6}$, for constant and
random scores (Lemma~\ref{lem:wellposed}); both bounds of Lemma~\ref{lem:sandwich} hold,
with equality on the diagonal; $t\lambda_t\le e^{B}$, and with constant scores
$t\lambda_t\ge a_0$ for $\rho$ up to $0.9$ and $r_i\le1/(a_0(i-j+1))$
(Lemmas~\ref{lem:lambda-upper}, \ref{lem:alpha0}, \ref{lem:ratio}); and
$\max_jP_{tj}\le(e^{2B}/t)^{1-\rho}$ for $B\in\{0,1,3\}$ (Theorem~\ref{thm:dilution}).
In \texttt{test\_potential\_framework}: row terms leave softmax and the fixed point
unchanged while a column term does not (Proposition~\ref{prop:rowshift}); the far mass of
a linear bias is bounded uniformly in the length, GRAPE-A, FoX and GRAPE-AP decay at least
linearly, and CABLE's stopped clock is a counterexample (Lemma~\ref{lem:farmass},
Proposition~\ref{prop:arch}); and both bounds of Corollary~\ref{cor:boundary}, as well as
those on $Z_t$ and $\phi_\delta$ in its proof, hold for $\delta$ from $10^{-3}$ to $1$,
$L\le512$ and $t\le10^5$; at $c=1.5$ and $t=10^4$ the upper bound in (ii) is within 1\% of
the window mass.

\subsection{Comparison with the limits}
\label{app:num-asymp}

Table~\ref{tab:num-asymp} compares finite lengths with the limits as $t\to\infty$; the
tolerances describe the tested settings, not uniform error bounds.

\begin{table}[ht]
\centering
\small
\caption{Limit statements against finite lengths. Tests in
\texttt{test\_unbalanced\_scaling}, except Theorem~\ref{thm:lambda}
(\texttt{test\_unbalanced\_recursion}) and the last row
(\texttt{test\_potential\_framework}).}
\label{tab:num-asymp}
\begin{tabular}{p{0.2\linewidth}p{0.3\linewidth}p{0.42\linewidth}}
\toprule
Result & Limit & Finite length \\
\midrule
Thm.~\ref{thm:lambda} & $t\lambda_t\to\astar$ & decreases towards $\astar$; about $10^{-3}$
above it at $t=2048$ \\
Thm.~\ref{thm:shape} & $tP_{t,\lfloor st\rfloor}\to f(s)$ & within $2\%$ at $n=1024$, error
halves from $n=256$ to $1024$ \\
Thm.~\ref{thm:near} & discrete profile, then $d^{-\rho}$ & within $1\%$ for $d\le8$ at
$n=1024$ \\
Remark~\ref{rem:window} & window mass $\approx(L/t)^{1-\rho}/\Gamma(2-\rho)$ & $L=64$:
doubling $n$ from $512$ to $1024$ multiplies it by $2^{-(1-\rho)}$ to $2\%$; within $5\%$
of the limit \\
Thm.~\ref{thm:iters} & $tP^{(k)}_{t,\lfloor st\rfloor}\to f(s)$, $k\ge2$ & $k=2$: ratio to
$f(s)$ constant in $s$ to $1\%$ at $n=1024$ (the amplitude: Appendix~\ref{app:num-obs}) \\
Prop.~\ref{prop:iid} & $tP\kappa/(Kf)\to1$ & binned mean within $2\%$ of one \\
Prop.~\ref{prop:rates} & near mass like $t^{-(1-c)}$, $1/\log t$, or $\to S_L/\zeta(c)$ &
ratio across lengths within $10\%$ of the rate \\
\bottomrule
\end{tabular}
\end{table}

For constant scores, $t\lambda_t$ at $t=128,512,2048$ is $0.8449,0.8410,0.8398$
($\astar=0.8393$) for $\rho=0.25$, $0.6471,0.6398,0.6375$ ($0.6366$) for $\rho=0.5$, and
$0.3833,0.3743,0.3715$ ($0.3704$) for $\rho=0.75$; longer single computations at $n=4096$
are closer still. The limit shape: $tP_{t,\lfloor st\rfloor}$ is compared with $f(s)$ for
$\rho\in\{0.25,0.5,0.75\}$ and $s\in\{0.05,0.3,0.7,0.95\}$, for example $0.515$ against
$0.514$ at $\rho=0.5$, $s=0.3$. Near the diagonal ($\rho=0.5$), $t^{1-\rho}P_{t,t-d}$ at
$d=0,1,2,8,64$ is $0.7982,0.4933,0.3809,0.2008,0.0706$, against
$0.7979,0.4931,0.3808,0.2008,0.0709$ from the discrete profile and $0.0705$ from the power
law at $d=64$. With i.i.d.\ scores uniform on $[-B,B]$, the mean of $tP\kappa/(Kf)$ over
the last 64 rows lies in $[0.996,1.017]$ for $\rho\in\{0.5,0.75\}$, $B\in\{1,3\}$ and $s$
near $0.05,0.3,0.7,0.9$.

\subsection{Effects the theory does not quantify}
\label{app:num-obs}

The theorems give limits, not rates; the following are measured, not proved.

\paragraph{Finite iterations.}
With $k=2$ the main deviation in the tested settings is the amplitude: $tP/f$ is nearly
constant in $s$ but not one. It is $1.20$, $1.13$ and $1.09$ at $n=1024,4096,16384$ for
$\rho=0.75$, $1.76$ at $n=4096$ for $\rho=0.9$, and within $1.3\%$ of one at $n=4096$ for
$\rho=0.5$. With $k=3$ the error is mainly a tilt in $s$, of $4$--$15\%$ for $\rho\ge0.75$
at $n=4096$--$16384$ (\path{scripts/finite_iteration_rate.py}).

\paragraph{The fixed point.}
The error of $t\lambda_t$ above shrinks roughly like $t^{-0.9}$ over the tested lengths for
all three values of $\rho$. This is the approach to the limit in $t$, not the convergence
of the iteration in $k$.

\paragraph{i.i.d.\ content.}
At $n=4096$ the standard deviation of $tP\kappa/(Kf)$ over the last 64 rows grows with $B$
and $\rho$, from $0.015$ to $0.098$: a mean close to one coexists with appreciable
row-to-row variation.

\section{Experimental details and supplementary results}
\label{app:exp}

\subsection{Setup}
\label{app:exp-setup}

\paragraph{Data and training.}
FineWeb-Edu~\citep{penedo2024fineweb} \texttt{sample-10BT} at revision \texttt{87f09149},
tokenized with the GPT-2 tokenizer, documents separated by end-of-text tokens. Validation
uses shard 013 (20M tokens, hash prefix \texttt{318c97404e7d7482}); all file hashes are
recorded in the data manifest. Models are decoder-only pre-norm Transformers with tied
embeddings, trained with AdamW, warmup and cosine decay under bf16 autocast
(Table~\ref{tab:hparams}); arms differ only in their position encoding and normalizer
(Table~\ref{tab:arms}). The OT arms use a fused Triton kernel for the prefix iteration,
with $k=13$ ($\rho=0.5$) and $k=7$ ($\rho=0.25$) iterations chosen so that the fixed-point
residual is at most $10^{-4}$; each run re-measured the residual after training. Gradients
flow through the $k$ unrolled iterations, not through the fixed point. Per head, these
iterations cost $\mathcal O(kn^2)$ work at length $n$. The kernel stores prefix carries at
row-block boundaries and recomputes within blocks during backward; at fixed block size the
saved state remains $\mathcal O(kn^2)$. The learnable log-distance exponents are
$c_h=\softplus(\theta_h)$, initialized log-spaced in $[0.1,1]$ over the heads; P6 and LC
add a linear gate of the layer input to $\theta_h$, and KERPLE's $p_h$, $a_h$ start uniform
in $(0,2)$ and $(0,1)$ as in its code.

\paragraph{Evaluation.}
Losses at different lengths average over the same initial predicted tokens of the
validation stream: 2M in the main 100M evaluations and in the interventions, 262{,}144 in
the far-mass probe. The far mass evaluates two validation sequences per length ($2N$ tokens)
because computing full attention couplings across all layers and heads at lengths up to
$16L$ is memory-intensive; with $N/2$ queries per sequence evaluated across 14 layers and
12 heads, each probe averages over hundreds of thousands of attention rows, yielding
negligible sequence-to-sequence variation ($<0.01$). For each query in the second half of
a sequence, the attention on keys more than one training length back is averaged over these
queries, layers and heads. YaRN-t is applied to trained checkpoints at evaluation time. For evaluation length $N$ and training length $L$, the
scaling factor is $s=\max(1,N/L)$. A RoPE frequency $\omega$ that makes $L\omega/(2\pi)$
rotations within the training window becomes $\omega/s$ up to one rotation and stays
$\omega$ from 32 rotations onward, with linear interpolation in the rotation count in
between. To adjust the attention temperature, queries and keys are both scaled by
$1+0.1\log s$, which multiplies their dot products by $(1+0.1\log s)^2$.

\begin{table}[!htbp]
\centering
\small
\caption{Training configurations; 32{,}768 tokens per step at both scales. Arms differ
only in their position encoding and normalizer; their settings are the overrides listed in
the accompanying configuration files (\texttt{configs/exp0002/arms.yaml}).}
\label{tab:hparams}
\begin{tabular}{lcc}
\toprule
 & Screening (25M) & Confirmation (100M) \\
\midrule
Width / depth / heads & 512 / 8 / 8 & 768 / 14 / 12 \\
MLP ratio, vocabulary & 4, 50{,}304 (tied) & 4, 50{,}304 (tied) \\
Non-embedding parameters & 25.2M & 99.2--99.5M \\
Training length & 512 & 1024 (also 2048, 4096) \\
Micro-batch $\times$ accumulation & $32\times2$ & $8\times4$ ($4\times4$, $2\times4$) \\
Tokens / steps & 0.5B / 15{,}258 & 2B / 61{,}035 \\
Learning rate & $3\times10^{-3}$ & $5.3\times10^{-4}$ (CABLE variant: $7.5\times10^{-4}$) \\
Warmup, schedule & 750 steps, cosine to $0.1\times$ & 1{,}220 steps, cosine to $0.1\times$ \\
Weight decay, gradient clip & 0.1, 1.0 & 0.1, 1.0 \\
Hardware & RTX 4090 & A100-SXM4-80GB, RTX 4090, DGX Spark \\
\bottomrule
\end{tabular}
\end{table}

\iftmlrreview
\begin{table}[!tp]
\else
\begin{table}[!t]
\fi
\centering
\caption{Positional schemes, with $d=t-j$. ``Bounded far mass'' is the guarantee of
Proposition~\ref{prop:arch} under bounded inputs, where the conditional cases are stated in
full. $c_h(x_t)$ is the positive exponent predicted from the layer input $x_t$ at query
position $t$. $S_t$ accumulates learned tokenwise clock increments, and $g_t$ is CABLE's
positive query-dependent weight (head indices suppressed).
As in their papers, GRAPE-A, GRAPE-AP, FoX and CABLE are used without RoPE; so is P5.
Every other arm with a bias adds it to RoPE.}
\label{tab:arms}
\footnotesize
\setlength{\tabcolsep}{4pt}
\begin{tabular}{lp{0.52\linewidth}l}
\toprule
Arm & Position term & Bounded far mass \\
\midrule
B0 & RoPE, no bias & no (dilutes) \\
P1 & RoPE $+$ learnable $-c_h\log(1+d)$, one exponent per head & iff $c_h>1$ \\
P2 & as P1 with fixed exponents (log-spaced $0.1$--$1$) & no (dilutes) \\
P3 & RoPE $+$ KERPLE log kernel $-p_h\log(1+a_hd)$~\citep{chi2022kerple} & iff $p_h>1$ \\
P4 (ALiBi) & RoPE $+$ $-m_hd$~\citep{press2022alibi} & yes \\
P5 & P1 without RoPE & iff $c_h>1$ \\
P6 & RoPE $+$ $-c_h(x_t)\log(1+d)$, exponent gated by the query's input & if $\inf_xc_h(x)>1$ \\
LC & as P6 on a content clock, $-c_h(x_t)\log(1+S_t-S_j)$ & if $\inf_xc_h(x)>1$ \\
GRAPE-M & multiplicative group action (RoPE-like) & no (dilutes) \\
GRAPE-A & additive group action, gated linear in $d$ & yes \\
GRAPE-AP & path-integral bias & yes \\
FoX & cumulative log forget gate & yes \\
CABLE & content clock with per-query weight, $g_t(S_t-S_j)$ & if the clock ticks at a positive rate \\
O1 & RoPE with the OT operator, $\rho=0.5$ & no (dilutes) \\
O2 & ALiBi with the OT operator, $\rho=0.5$ & yes (fixed point) \\
\bottomrule
\end{tabular}
\end{table}

\FloatBarrier
\subsection{Preregistered tests}
\label{app:exp-prereg}

Table~\ref{tab:prereg} lists the four preregistered experiments with their tests and
outcomes. Loss tests compare a mean paired difference (same seed or same checkpoint) with a
threshold and require the same sign on every seed or checkpoint. In the encoding comparison
the threshold is $\delta_e=\max(2\sigma_e,0.01)=0.117$, where $\sigma_e$ is the seed
standard deviation of B0's $L_{\mathrm{ext}}$; the main test also requires CABLE to beat B0
by $\delta_e$ and to stay within $0.01$ of B0's $1\times$ loss. Because $\delta_e$ far
exceeds the seed standard deviation of the paired differences ($0.0013$--$0.0027$), the
retests use the threshold $0.01$ on new seeds and are reported separately. For the
test-time interventions, the threshold applies to the mean over all checkpoints of one
intervention type, pooled across arms and scales; Table~\ref{tab:tail} splits these sets by
scale only for reporting, so the 100M LC repair mean ($-0.009$) is not tested on its own.

\begin{table}[!htbp]
\centering
\caption{Preregistered tests and their outcomes. Loss thresholds apply to mean paired
gains; ``every seed'' and ``every checkpoint'' require consistent signs. Far-mass
thresholds apply to seed means; all thresholds were fixed before the runs.}
\label{tab:prereg}
\footnotesize
\begin{tabular}{lll}
\toprule
Test & Observed & Outcome \\
\midrule
\multicolumn{3}{l}{\emph{OT versus softmax (100M)}} \\
best $\rho$ below softmax at $L$ by $0.0100$, every seed & $0.0095$ & fails \\
best $\rho$ below softmax / log-$n$ at $2L$ by $0.057$, every seed & $0.455$, $0.303$ & holds \\
\addlinespace
\multicolumn{3}{l}{\emph{Encoding comparison (25M screening, 100M confirmation)}} \\
CABLE below B0 + YaRN-t in $L_{\mathrm{ext}}$ by $\delta_e=0.117$, every seed & $0.058$ &
fails \\
GRAPE-M or GRAPE-AP below CABLE by $\delta_e$, every seed & no & fails \\
O2 below CABLE and ALiBi by $\delta_e$, every seed & no & fails \\
\addlinespace
\multicolumn{3}{l}{\emph{Retest and tail interventions (100M retest; 25M and 100M checkpoints)}} \\
CABLE below B0 + YaRN-t in $L_{\mathrm{ext}}$ by $0.01$, every seed, $L=1024$, $2048$ &
$0.059$, $0.065$ & holds \\
P1 with $c>1$ below $c<1$ at $32\times$ by $0.01$, every seed & $0.488$--$0.678$ & holds \\
\quad and far-mass growth $4\times\to16\times$: $c>1$ $\le0.02$, $c<1$ $>0.1$ & $0.044$,
$0.205$ & fails \\
cap, repair, window: every checkpoint, pooled mean above $0.01$ ($0.1$) & $30/30$ &
holds \\
\addlinespace
\multicolumn{3}{l}{\emph{Retest at $L=4096$ (100M) and exponent margins (25M)}} \\
as the retest above, at $L=4096$ & $0.074$ & holds \\
$\delta_0=0.5$: far-mass growth $\le0.02$; below $\delta_0=0$ at $32\times$ by $0.01$ &
$0.004$, $0.042$ & holds \\
far-mass growth ordered $\delta_0=0>0.25>0.5$, every seed & $3/3$ & holds \\
\bottomrule
\end{tabular}
\end{table}

\FloatBarrier
\subsection{Supplementary results}
\label{app:exp-results}

\paragraph{OT mechanism (exploratory).}
\label{app:exp-ot}\label{app:exp-yarn}
In the 25M screening models, the distance part explains a layer-averaged 11--27\% of the
variance of the OT correction across $\rho\in\{0.1,0.25,0.5,0.75\}$, and 18--19\% at
$\rho=0.5$. To remove the operator's built-in distance decay, the uniform-target operator
replaces the target column mass $1$ by $T_{tj}=\sum_{i=j}^{t}1/i$, the mass that uniform
attention delivers, that is, $\log M^{(k)}_{tj}$ by $\log(M^{(k)}_{tj}/T_{tj})$
in~\eqref{eq:iteration}; its fixed point is $P_{tj}=\lambda_tK_{tj}(M_{tj}/T_{tj})^{-\gamma}$,
and with constant scores the correction vanishes. Its $2L$ loss is above that of both the
original target and softmax, and adding it to the log-distance bias worsens both lengths
(Table~\ref{tab:ot-ablation}a). Since the target also changes the fixed point, this does
not isolate the content part. At 100M, applying YaRN-t to both softmax and OT reduces their
loss gap at $2L$ by $97.7\%$ (Table~\ref{tab:ot-ablation}b). A single 100M $\rho=0.5$ run
at half the learning rate is worse at both lengths; this comparison tests only that lower
rate. The prefix operator also reduces attention sinks, the mass on the first
token~\citep{xiao2024streaming}; we did not find this to matter for extrapolation.
Across evaluation lengths from $L$ to $8L$ (512 to 4096), the post-training fixed-point
residual remains bounded below $7.4\times 10^{-5}$ for $\rho\le 0.5$ and $3.0\times 10^{-4}$
for $\rho=0.75$, confirming that extrapolation degradation does not stem from fixed-point
solver divergence.

\begin{table}[!htbp]
\centering
\caption{Exploratory OT ablations, validation loss in nats per token. (a) Screening
configuration (25M, $L=512$, $\rho=0.5$, $k=13$), means over two seeds. (b) The 100M
checkpoints of Table~\ref{tab:exp0001} at $2L=2048$ on about 1M predictions, means over
three seeds; the gap is softmax minus OT.}
\label{tab:ot-ablation}
\footnotesize
\begin{minipage}[t]{0.46\linewidth}
\centering
(a) Uniform target, 25M\\[3pt]
\begin{tabular}{lcc}
\toprule
Arm & $L$ & $2L$ \\
\midrule
softmax & 3.471 & 3.635 \\
OT & 3.468 & 3.483 \\
OT, uniform target & 3.481 & 3.968 \\
log-distance bias & 3.461 & 3.423 \\
\quad $+$ OT, uniform target & 3.469 & 3.485 \\
\bottomrule
\end{tabular}
\end{minipage}\hfill
\begin{minipage}[t]{0.5\linewidth}
\centering
(b) Evaluation-time scaling, 100M\\[3pt]
\begin{tabular}{lccc}
\toprule
RoPE frequencies & softmax & OT & gap \\
\midrule
original & 3.6113 & 3.1539 & 0.457 \\
interpolation only & 3.0661 & 3.0349 & 0.031 \\
YaRN-t & 3.0379 & 3.0273 & 0.011 \\
\bottomrule
\end{tabular}
\end{minipage}
\end{table}

\paragraph{Screening.}
\label{app:exp-screen}
Table~\ref{tab:screen} gives the screening results by arm. B0 with YaRN-t reaches
$L_{\mathrm{ext}}=3.461$. With the same bias, adding OT is slightly worse (O2 against
ALiBi: $+0.005$ and $+0.004$ on the two seeds).

\iftmlrreview
\begin{table}[!tp]
\else
\begin{table}[!t]
\fi
\centering
\caption{Screening at 25M, $L=512$, means over two seeds, grouped by the cases of
Proposition~\ref{prop:arch}. Loss in nats per token over 2M predictions; far mass from the
probe of Section~\ref{sec:exp}; last column, the share of learned exponents above one, over
heads, or heads $\times$ queries for P6 and LC.}
\label{tab:screen}
\small
\setlength{\tabcolsep}{4pt}
\begin{tabular}{llccccc}
\toprule
Case & Arm & $1\times$ & $L_{\mathrm{ext}}$ & far mass $4\times$ & $16\times$ & $c>1$ \\
\midrule
dilutes & B0 & 3.474 & 3.865 & 0.679 & 0.889 & -- \\
 & GRAPE-M & 3.471 & 3.930 & 0.701 & 0.899 & -- \\
 & O1 & 3.469 & 3.573 & 0.428 & 0.646 & -- \\
 & P2 & 3.461 & 3.460 & 0.424 & 0.627 & 0.00 \\
\addlinespace
learned & P5 & 3.489 & 3.498 & 0.253 & 0.480 & 0.34 \\
 & P6 & 3.460 & 3.440 & 0.364 & 0.561 & 0.18 \\
 & P1 & 3.461 & 3.432 & 0.370 & 0.551 & 0.15 \\
 & P3 & 3.463 & 3.413 & 0.192 & 0.279 & 0.64 \\
 & LC & 3.456 & 3.403 & 0.221 & 0.309 & 0.56 \\
 & CABLE & 3.467 & 3.400 & 0.089 & 0.050 & -- \\
\addlinespace
bounded & ALiBi (P4) & 3.465 & 3.408 & 0.030 & 0.024 & -- \\
 & GRAPE-A & 3.475 & 3.417 & 0.013 & 0.010 & -- \\
 & GRAPE-AP & 3.485 & 3.428 & 0.023 & 0.020 & -- \\
 & FoX & 3.490 & 3.429 & 0.015 & 0.008 & -- \\
 & O2 & 3.469 & 3.413 & 0.032 & 0.024 & -- \\
\bottomrule
\end{tabular}
\end{table}

\paragraph{Confirmation and retests at 100M.}
\label{app:exp-confirm}
The confirmation trains eight arms with three seeds each (Table~\ref{tab:confirm}): B0,
log-$n$ softmax, ALiBi, CABLE and LC (screening ranks 1 and 2; LC exploratory), GRAPE-M,
GRAPE-AP and O2; B0 + YaRN-t is evaluated on the B0 checkpoints. CABLE was trained at two
learning rates, and a preregistered rule selected the one with lower $1\times$ loss. CABLE
beats B0 + YaRN-t in $L_{\mathrm{ext}}$ by $0.058$ ($0.059$, $0.060$, $0.056$ per seed),
below $\delta_e=0.117$, with a $1\times$ loss $0.017$ lower than B0. Both GRAPE variants and
O2 have higher $L_{\mathrm{ext}}$ than CABLE, and O2 is above ALiBi on every seed (by
$0.0031$--$0.0047$). Table~\ref{tab:retest-abs} gives the absolute losses of the retests.

\begin{table}[!htbp]
\centering
\caption{Confirmation at 100M: validation loss in nats per token (mean $\pm$ sd over three
seeds). YaRN-t is defined only for arms with RoPE. CABLE uses the learning-rate variant
selected by the preregistered rule.}
\label{tab:confirm}
\small
\begin{tabular}{lccccc}
\toprule
Arm & $1\times$ & $L_{\mathrm{ext}}$ & $8\times$ & $L_{\mathrm{ext}}$, YaRN-t & ms/step \\
\midrule
B0 (RoPE) & $3.0440\pm0.0019$ & $4.1136\pm0.0584$ & $5.468\pm0.106$ & $3.0397\pm0.0016$ & 257 \\
log-$n$ softmax & $3.0464\pm0.0038$ & $3.9505\pm0.0827$ & $5.369\pm0.339$ & $3.0541\pm0.0054$ & 347 \\
GRAPE-M & $3.0433\pm0.0014$ & $4.1982\pm0.1076$ & $5.576\pm0.183$ & -- & 262 \\
GRAPE-AP & $3.0418\pm0.0008$ & $3.0105\pm0.0013$ & $3.004\pm0.001$ & -- & 482 \\
ALiBi (P4) & $3.0312\pm0.0007$ & $2.9932\pm0.0006$ & $2.979\pm0.001$ & $3.0585\pm0.0019$ & 260 \\
O2 (P4 + OT) & $3.0351\pm0.0011$ & $2.9972\pm0.0011$ & $2.983\pm0.001$ & $3.0458\pm0.0025$ & 1582 \\
LC & $\mathbf{3.0220\pm0.0021}$ & $2.9912\pm0.0016$ & $3.028\pm0.009$ & $2.9961\pm0.0027$ & 341 \\
CABLE & $3.0269\pm0.0022$ & $\mathbf{2.9814\pm0.0023}$ & $\mathbf{2.962\pm0.002}$ & -- & 287 \\
CABLE, B0's lr & $3.0270\pm0.0009$ & $2.9818\pm0.0011$ & $2.962\pm0.002$ & -- & 287 \\
\bottomrule
\end{tabular}
\end{table}

\begin{table}[!htbp]
\centering
\caption{Retest at 100M on new seeds: validation loss in nats per token (mean $\pm$ sd) at
the training length ($1\times$) and $L_{\mathrm{ext}}$.}
\label{tab:retest-abs}
\footnotesize
\begin{tabular}{lcccc}
\toprule
 & \multicolumn{2}{c}{$1\times$} & \multicolumn{2}{c}{$L_{\mathrm{ext}}$} \\
\cmidrule(lr){2-3}\cmidrule(lr){4-5}
Length (seeds) & B0 & CABLE & B0 + YaRN-t & CABLE \\
\midrule
1024 (5) & $3.0445\pm0.0039$ & $3.0258\pm0.0012$ & $3.0391\pm0.0039$ & $2.9801\pm0.0012$ \\
2048 (3) & $3.0233\pm0.0028$ & $2.9932\pm0.0012$ & $3.0302\pm0.0039$ & $2.9649\pm0.0012$ \\
4096 (3) & $3.0201\pm0.0045$ & $2.9749\pm0.0009$ & $3.0343\pm0.0038$ & $2.9602\pm0.0009$ \\
\bottomrule
\end{tabular}
\end{table}

\paragraph{Test-time interventions.}
\label{app:exp-test}
The interventions act on the total additive bias of each head after loading a trained
checkpoint and are applied in both the reference and the fused attention;
Table~\ref{tab:tail} lists the checkpoints, and Figure~\ref{fig:intervention-examples}
shows one 25M checkpoint per intervention. LC repair starts from its original
content-clock bias at $d=L$, then extends the tail in token distance with exponent $1.5$.
Both LC checkpoints show lower far mass and loss at $8L$, as predicted for stronger tail
suppression; the smaller loss reduction accompanies LC's lower initial far mass, and the
fixed exponent can also weaken decay in heads with learned exponents above $1.5$.
While a clock-consistent repair extending learned clock increments rather than token distance
could preserve token-rate sensitivity, LC's pre-existing decay ($0.36$ far mass)
leaves less room for improvement than the diluting arms. An exploratory repair of the
three 100M token-distance log-bias checkpoints (Table~\ref{tab:exp0001}, last row) changes
loss in the same direction.

\begin{figure}[!ht]
\centering
\includegraphics[width=\linewidth]{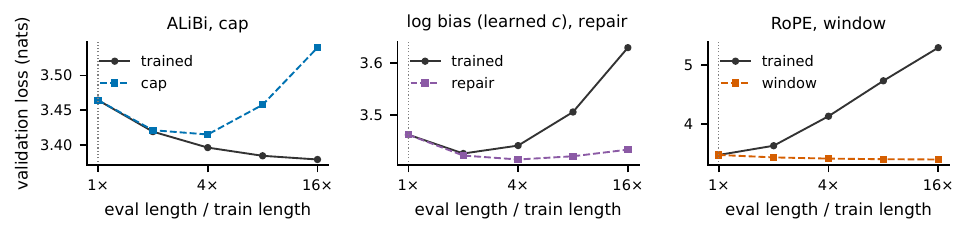}
\caption{Loss against the evaluation length for one 25M checkpoint per intervention, before
(dark grey) and after; the curves coincide up to the training length by construction.}
\label{fig:intervention-examples}
\end{figure}

\begin{table}[!ht]
\centering
\caption{Test-time interventions beyond the training length
(Figure~\ref{fig:interventions}). Loss change in nats at $16\times$ (25M) or $8\times$
(100M), mean and range over checkpoints; far mass at the same length before and after.
The $1\times$ losses agree within $10^{-4}$ in every case.}
\label{tab:tail}
\footnotesize
\setlength{\tabcolsep}{3pt}
\begin{tabular}{l>{\raggedright\arraybackslash}p{0.36\linewidth}cccc}
\toprule
 & Checkpoints (seeds each) & Scale & Loss change & \multicolumn{2}{c}{Far mass} \\
\cmidrule(lr){5-6}
 & & & & before & after \\
\midrule
cap & ALiBi, GRAPE-A, FoX, CABLE (2) & 25M & $+0.060$ ($+0.008$ to $+0.161$) &
0.007--0.058 & 0.036--0.201 \\
cap & ALiBi, CABLE, CABLE (default lr) (3) & 100M & $+0.020$ ($+0.012$ to $+0.064$) &
0.003--0.045 & 0.036--0.093 \\
\addlinespace
repair & P1, P2, P6 (2) & 25M & $-0.264$ ($-0.312$ to $-0.198$) & 0.55--0.63 & 0.32--0.39 \\
repair & LC (2) & 100M & $-0.009$ ($-0.011$ to $-0.008$) & 0.36--0.37 & 0.26--0.27 \\
repair & log bias, exploratory (3) & 100M & $-0.182$ ($-0.210$ to $-0.166$) & 0.58--0.61 &
0.39--0.44 \\
\addlinespace
window & B0 (2) & 25M & $-1.93$ ($-1.96$ to $-1.90$) & 0.89 & 0.03 \\
window & B0 (3) & 100M & $-2.46$ ($-2.55$ to $-2.34$) & 0.87--0.88 & 0.01 \\
\bottomrule
\end{tabular}
\end{table}

\paragraph{Training-time constraints.}
\label{app:exp-train}
Table~\ref{tab:constraints} reports both training-time experiments. The first constrains
the P1 exponents to $c=1+\softplus(\theta)>1$ or $c=\sigma(\theta)<1$; the second imposes
$c\ge1+\delta_0$. Under $c>1$, many heads settle just above one (per-seed minima
$1.0001$--$1.0002$); their transition scales exceed $e^{10}$, so they remain in regime~(i)
of Corollary~\ref{cor:boundary} through $32L$. The margin follows from a window, a length
range and a tolerance; the indicator $\delta_0\log n\approx1$ alone does not decide whether
a tolerance is met. For the prior $(1+d)^{-c}$ with all heads at $c=1+\delta_0$, $L=512$ and
the far mass averaged as in Section~\ref{sec:exp}, the growth from $4L$ to $16L$ is
$0.088$, $0.043$ and $0.010$ for $\delta_0=0.1$, $0.25$, $0.5$, so of these only $0.5$
meets the tolerance $0.02$, although $\delta_0\log16L\approx1$ already at $\delta_0=0.1$.

\begin{table}[!htbp]
\centering
\caption{Training-time constraints on the P1 exponents at 25M, $L=512$, means over three
seeds: validation loss, far mass at $16\times$, its growth from $4\times$ to $16\times$,
measured and predicted by the prior with the learned exponents (without content scores;
post hoc in the upper block), and the share of heads at the bound ($c<1.1$ in the upper
block, within $0.01$ of $1+\delta_0$ in the lower). The lower block extends
Table~\ref{tab:margin}.}
\label{tab:constraints}
\footnotesize
\begin{tabular}{lcccccc}
\toprule
 & & & & \multicolumn{2}{c}{Far-mass growth} & \\
\cmidrule(lr){5-6}
Arm & $1\times$ & $32\times$ & Far mass $16\times$ & measured & predicted & Heads at bound \\
\midrule
\multicolumn{7}{l}{\emph{Constraint $c>1$ (seeds 0--2)}} \\
P1, $c<1$ & 3.438 & 4.005 & 0.585 & 0.205 & 0.254 & -- \\
P1, free & 3.439 & 3.888 & 0.554 & 0.197 & 0.234 & -- \\
P1, $c>1$ & 3.444 & 3.430 & 0.163 & 0.044 & 0.067 & 39--44\% \\
ALiBi (P4) & 3.443 & 3.359 & 0.023 & $-0.006$ & 0.000 & -- \\
\addlinespace
\multicolumn{7}{l}{\emph{Margin $c\ge1+\delta_0$ (seeds 3--5)}} \\
P1, $\delta_0=0$ & 3.444 & 3.427 & 0.160 & 0.032 & 0.067 & 23--30\% \\
P1, $\delta_0=0.25$ & 3.445 & 3.395 & 0.113 & 0.015 & 0.023 & 25--28\% \\
P1, $\delta_0=0.5$ & 3.448 & 3.385 & 0.071 & 0.004 & 0.006 & 27--31\% \\
ALiBi (P4) & 3.441 & 3.355 & 0.023 & $-0.006$ & -- & -- \\
\bottomrule
\end{tabular}
\end{table}

\paragraph{Far mass and distant use at 100M (exploratory).}
\label{app:exp-distant}
Table~\ref{tab:distant} gives the far mass and two distant-use probes on the confirmation
checkpoints. From $4\times$ to $16\times$ the loss of B0 and GRAPE-M rises by
$1.35$--$1.41$, while the arms with a bounded far mass keep a flat loss; LC is in between
($+0.38$, with 44\% of per-query exponents above one). Retrieval places a
passkey~\citep{mohtashami2023landmark} or a needle~\citep{kamradt2023needle} at five depths, with
20 trials per depth. At $4\times$ the other arms succeed only at the depth closest to the
query, if at all, whereas CABLE's passkey accuracy is $0.62$ even at the farthest depth,
about $3{,}700$ tokens back (needle: $0.05$). On the repetitive passkey filler, CABLE's
content increments $\Delta S_l = \mathrm{ReLU}(w_c x_l + b_c)$ vanish, so its clock
effectively pauses and measures semantic rather than token distance: its far mass is
$0.19$--$0.20$ at $8\times$, and its attention to the key span is $0.04$ at $4\times$
(others $\le0.003$). In every CABLE checkpoint one to three heads of layers 5--6 put
$0.6$--$1.0$ of their attention on the passkey. Conversely, on natural text needles where
the clock ticks continuously, accuracy drops to $0.05$ at the farthest depth, showing that
this retrieval relies on low-information fillers rather than general distant access. Copying
gain~\citep{olsson2022incontext} is the loss reduction on the final 128-token target when
its copy replaces a fragment $\Delta$ tokens earlier, on $5L$-token sequences with 64
shared trials; a shuffled copy gains at most $0.02$. CABLE's gains hold within $0.006$ on
five new seeds and are $0.17$ and $0.07$ at $\Delta=2L$, $4L$ for training length 2048.

\begin{table}[!htbp]
\centering
\caption{Exploratory far mass and distant use at 100M, $L=1024$, means over three seeds
(two for LC's far mass and copying gain due to an incomplete seed-0 backup; see
Section~\ref{app:exp-registry}). Retrieval accuracy is averaged over five key depths;
copying gain in nats at distance $\Delta$ ($N=5L$). Un-fused reference arms at $16\times$
(16k) exceed the 48GB GPU memory cap. CABLE uses the selected learning-rate variant
($7.5\times 10^{-4}$).}
\label{tab:distant}
\footnotesize
\setlength{\tabcolsep}{3pt}
\begin{tabular}{lccccccccccc}
\toprule
 & \multicolumn{3}{c}{Far mass} & \multicolumn{2}{c}{Passkey} & \multicolumn{2}{c}{Needle}
 & \multicolumn{4}{c}{Copying gain, $\Delta=$} \\
\cmidrule(lr){2-4}\cmidrule(lr){5-6}\cmidrule(lr){7-8}\cmidrule(lr){9-12}
Arm & $2\times$ & $4\times$ & $8\times$ & $2\times$ & $4\times$ & $2\times$ & $4\times$ & $L/2$ & $L$ &
$2L$ & $4L$ \\
\midrule
B0 (RoPE) & 0.458 & 0.756 & 0.876 & 0.00 & 0.00 & 0.00 & 0.00 & 0.004 & 0.007 & 0.010 & 0.014 \\
log-$n$ softmax & 0.413 & 0.678 & 0.812 & 0.00 & 0.00 & 0.00 & 0.00 & 0.043 & 0.021 & 0.013 & 0.016 \\
GRAPE-M & 0.474 & 0.772 & 0.887 & 0.00 & 0.00 & 0.00 & 0.00 & 0.003 & 0.006 & 0.021 & 0.044 \\
GRAPE-AP & 0.014 & 0.002 & 0.003 & 0.19 & 0.00 & 0.20 & 0.07 & 1.213 & 0.327 & 0.007 & 0.000 \\
ALiBi (P4) & 0.006 & 0.003 & 0.003 & 0.25 & 0.15 & 0.25 & 0.10 & 1.453 & 0.079 & 0.000 & 0.000 \\
O2 (P4 + OT) & 0.008 & 0.005 & 0.004 & 0.27 & 0.14 & 0.25 & 0.09 & 1.381 & 0.116 & 0.001 & 0.000 \\
LC & 0.182 & 0.271 & 0.365 & 0.33 & 0.00 & 0.21 & 0.04 & 1.129 & 0.116 & 0.009 & 0.002 \\
CABLE & 0.072 & 0.066 & 0.039 & 0.94 & 0.54 & 0.46 & 0.23 & 1.518 & 0.881 & 0.256 & 0.107 \\
\bottomrule
\end{tabular}
\end{table}

\FloatBarrier
\subsection{Compute and code}
\label{app:exp-registry}

\iftmlrreview
\else
Public code:
\url{https://github.com/homw314159-spec/causal-ot-attention},
tag \href{https://github.com/homw314159-spec/causal-ot-attention/tree/v0.1.0}{\texttt{v0.1.0}}
(commit \href{https://github.com/homw314159-spec/causal-ot-attention/commit/e60a38f72276f05c2b216f96c4ddb33b9a75b4e4}{\texttt{e60a38f}}).
The public release has a fresh history; the reported runs record clean development
commits and were not rerun for this release. The file \texttt{configs/reproduction.yaml}
provides training recipes; \texttt{docs/experiments.md} maps the protocols and probes to
scripts. It summarizes the preregistrations rather than publishing their original change
logs. Development history, data and checkpoints are not distributed.
\fi
Table~\ref{tab:runs} lists the training runs by experiment, with the identifiers used in
the code.

\begin{table}[!htbp]
\centering
\caption{Training runs. Confirmation runs lost with the first machine were retrained and
reproduce their $1\times$ losses to $0.0011$. In the retest at 4096, both arms of a seed ran
on the same machine.}
\label{tab:runs}
\footnotesize
\begin{tabular}{lllrl}
\toprule
Experiment & Id & Part & Runs & Hardware \\
\midrule
OT versus softmax & 0001 & main & 12 & A100-SXM4-80GB \\
 & & half learning rate & 1 & A100-SXM4-80GB \\
 & & log-distance bias & 3 & A100-SXM4-80GB \\
Encoding comparison & 0002 & screening & 30 & RTX 4090 \\
 & & confirmation & 27 & two A100-SXM4-80GB machines \\
Retest and interventions & 0003 & retest & 16 & one four-GPU A100 machine \\
 & & training-time constraint & 12 & RTX 4090 \\
Retest at 4096 and margins & 0004 & retest at 4096 & 6 & RTX 4090 (2 seeds), DGX Spark (1) \\
 & & margins & 12 & RTX 4090 \\
\bottomrule
\end{tabular}
\end{table}

\end{document}